\documentclass[conference]{IEEEtran}
\usepackage{times}

\usepackage[numbers]{natbib}
\usepackage{multicol}
\usepackage[bookmarks=true]{hyperref}

\usepackage{xcolor}
\usepackage{amsmath}
\usepackage{amssymb}
\usepackage{mathtools} 
\usepackage{xcolor}
\usepackage{subcaption}
\usepackage{todonotes}

\usepackage{balance}

\usepackage{booktabs}    
\usepackage{multirow}    
\usepackage{tabularx}    
\usepackage{caption}     
\usepackage{algorithm}
\usepackage{algorithmic}

\usepackage[normalem]{ulem} 
\definecolor{revcolor}{RGB}{0,90,180}

\newcommand{\rev}[1]{{#1}}
\newcommand{\del}[1]{}

\begin{document}

\title{Guided Streaming Stochastic Interpolant Policy}

\newcommand{\traj}{\tau}
\newcommand{\past}{\traj_{0:t}}
\newcommand{\future}{\traj_{t+1:T}}
\newcommand{\futureprime}{\traj'_{t+1:T}}
\newcommand{\cost}{J}
\newcommand{\base}{p}
\newcommand{\target}{q}
\newcommand{\optimal}{\target^*}
\newcommand{\action}{a}
\newcommand{\velocity}{v}

\newcommand{\note}[1]{{\color{red}\textbf{Note: }#1}}

\newcommand{\mypara}[1]{\medskip\noindent\textbf{#1}~}

\author{\authorblockN{Puming Jiang$^{1, \ddagger}$,
Meiyi Wang$^{2, \ddagger}$,
Kelvin Lin$^{3}$, 
Ce Hao$^{4}$ and
Harold Soh$^{5}$}
\authorblockA{School of Computing, National University of Singapore\\
$^{\ddagger}$Equal contribution\\
Email: \{p.jiang$^{1}$, meiyi.wang$^{2}$, cehao$^{4}$\}@u.nus.edu, \{klinzw$^{3}$, harold$^{5}$\}@nus.edu.sg}
}

\maketitle

\begin{abstract}

Inference-time guidance is essential for steering generative robot policies toward dynamic objectives without retraining, yet existing methods are largely confined to chunk-based architectures that exhibit high latency and  lack the reactivity needed for test-time preference alignment or obstacle avoidance. In this work, we formally derive the optimal guidance term for Stochastic Interpolants (SI) by analyzing the value function's time evolution via the Backward Kolmogorov Equation, establishing a modified drift that theoretically guarantees sampling from a target distribution. We apply this framework to real-time control through the Streaming Stochastic Interpolant Policy (SSIP), which generalizes the deterministic Streaming Flow Policy (SFP). Unifying this guidance law with the streaming architecture enables fast and reactive control. To support diverse deployment needs, we propose two complementary mechanisms: training-free Stochastic Trajectory Ensemble Guidance (STEG) that computes gradients on-the-fly for zero-shot adaptation, and training-based Conditional Critic Guidance (CCG) for amortized inference. Empirical evaluations demonstrate that our guided streaming approach significantly outperforms conventional chunk-based policies in reactivity and provides superior, physically valid guidance for dynamic, unstructured environments.

\end{abstract}

\IEEEpeerreviewmaketitle

\section{Introduction}

Modern generative approaches, such as Diffusion Models~\cite{song2020score, ho2020denoising, chi2023diffusion} and Flow Matching~\cite{lipman2022flow, liu2022flow}, excel at capturing complex, multi-modal behaviors from demonstrations. However, their standard formulations do not explicitly account for the dynamic constraints of physical deployment. In real-world settings, a robot must not only imitate demonstrated motion but also adapt to safety constraints (e.g., obstacles) and evolving user preferences absent from the training data. Consequently, the utility of a policy is determined not only by its base capabilities, but also by how readily it can be steered at inference time toward additional objectives during execution, without the cost of retraining.

\begin{figure}[t]
    \centering
    \includegraphics[width=0.45\textwidth]{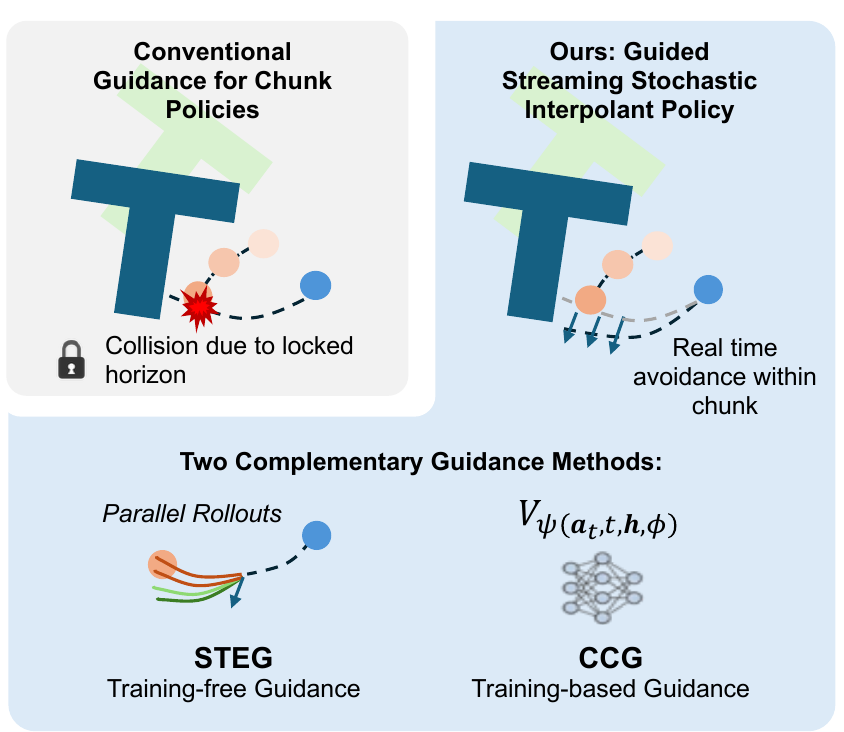}
    \caption{We contribute a principled inference-time guidance framework for streaming generative policies. It enables fast reactive behavior within the action chunk, and can be deployed in either a training-based or training-free manner.}
    \label{fig:teaser}
\end{figure}

This capability is essential for deploying generative policies in unstructured, real-world environments and is commonly referred to as \emph{inference-time guidance}. Originally introduced for controlled image generation~\cite{dhariwal2021diffusion}, inference-time guidance has since been adapted to robot control to enforce safety constraints~\cite{feng2024ltldog, deng2025safebimanual} and to align behavior with human preferences~\cite{wang2025inference}.
However, existing guidance methods are largely tailored to chunk-based policies, such as Diffusion Policy~\cite{chi2023diffusion}, and inherit their limitations. In particular, actions must be generated for an entire chunk before execution begins, introducing latency and often leading to discontinuous motion~\cite{jiang2025streaming}. Moreover, once a chunk is committed, execution proceeds open-loop over the chunk horizon, preventing adaptation to objectives that change mid-execution (e.g., moving obstacles). As a result, these methods are suboptimal for dynamic environments\rev{, as illustrated in Fig.~\ref{fig:teaser}}.

To address this challenge at a fundamental level, we ground our approach in the framework of \emph{Stochastic Interpolants} (SI)~\cite{albergo2023stochastic}. SI provides a unifying formulation that encompasses score-based diffusion models (SDEs)~\cite{song2020score} and flow matching methods (ODEs)~\cite{lipman2022flow, liu2022flow}. It has also been applied to robot policies to exploit the property that sampling need not originate from pure Gaussian noise, which accelerates the inference process~\cite{chen24bridger}. While guidance mechanisms are well established for standard diffusion models, a rigorous treatment within the more general SI framework remains underexplored.

In this work, we first derive the optimal inference-time guidance for Stochastic Interpolants. This formulation is \emph{general} and extends beyond policy guidance. We then propose the \emph{Streaming Stochastic Interpolant Policy (SSIP)}, a streaming generative control framework that addresses the latency and motion smoothness limitations of chunk-based policies~\cite{chen24bridger}. Similar to very recent work on streaming flow policies (SFP)~\cite{jiang2025streaming}, SSIP aligns generative evolution with the robot's physical execution time to enable incremental action generation during execution. However, unlike prior streaming approaches, SSIP instantiates the more general Stochastic Interpolant framework. This changes the dynamics from a deterministic Ordinary Differential Equation (ODE) to a Stochastic Differential Equation (SDE), rendering the streaming policy mathematically compatible with the optimal inference-time guidance derived for SI.

With the theoretical guidance formulation and the SSIP backbone established, we propose two complementary strategies for computing the guidance gradient:
\begin{itemize}
    \item {Stochastic Trajectory Ensemble Guidance (STEG):} A training-free approach utilizing parallel inference-time rollouts. By backpropagating through the differentiable trajectory ensemble, STEG \del{computes the optimal} \rev{estimates the} guidance gradient on-the-fly for zero-shot adaptation.
    \item {Conditional Critic Guidance (CCG):} A training-based approach that amortizes the utility gradient into a learned critic network, which offers efficient inference for known obstacle classes.
\end{itemize}
\del{Experiments in two domains (Push-T and Robomimic) show that our unified framework significantly improves latency and reactivity compared to existing Diffusion and Flow-Matching policies with guidance.}
\rev{Experiments on Push-T and Robomimic show that our unified framework improves latency and reactivity over existing guided Diffusion and Flow-Matching policies. For example, in the challenging Push-T Chase setting, chunk-based DP and FP achieve only 12.0\% and 8.0\% success, while SSIP variants maintain 30.0--36.0\%. We further validate the framework on a real Panda robot for moving-obstacle avoidance and position-preference grasping, demonstrating its applicability beyond simulation.}

\section{Background \& Related Work}

\mypara{Generative Robot Policies.}
Recent work has applied generative modeling to robot policy learning, commonly using action chunking to represent complex behaviors by predicting fixed-horizon action sequences (e.g., \(a_{t:t+H}\)). Representative examples include Diffusion Policy (DP)~\cite{chi2023diffusion}, Flow Policies (FP)~\cite{lipman2022flow}, and Action Chunking with Transformers (ACT)~\cite{zhao2023learning}.  Despite their expressiveness, these approaches require generating an entire action chunk prior to execution and operate open-loop within each chunk, with feedback incorporated only at chunk boundaries.
\rev{Recent methods such as Consistency Policy~\cite{prasad2024consistency}, One-Step Diffusion Policy~\cite{wang2025onestep}, and DPPO~\cite{ren2025diffusion} improve the speed or performance of chunk-based diffusion policies; using smaller chunks can also shorten the open-loop horizon. Real-time chunking methods such as RTC~\cite{black2026realtime} further improve asynchronous execution of chunk-based policies via inpainting and cross-chunk continuity. These directions are complementary to our focus on deriving test-time guidance for execution-time-aligned policies.} 

Very recent work on Streaming Flow Policies (SFP)~\cite{jiang2025streaming} removes the reliance on action chunking by modeling action trajectories directly as flow trajectories. By aligning generative flow time with physical execution time, SFP enables incremental action generation and reduces latency while preserving performance. \rev{Our work builds on this execution-time-aligned perspective and studies how to equip streaming policies with principled test-time guidance.}

\mypara{Stochastic Interpolants.}
Stochastic Interpolants (SI)~\cite{albergo2023stochastic} provide a unified framework that connects
flow-based generative models (probability-flow ODEs) and diffusion-based models (SDEs).
Given endpoint densities $\rho_0$ and $\rho_1$, SI defines a family of intermediate marginals
$\{\rho_t\}_{t\in[0,1]}$ via a stochastic interpolant
\begin{equation}
    \mathbf{x}_t = I(t,\mathbf{x}_0,\mathbf{x}_1,\mathbf{u}) + \gamma(t)\mathbf{z},\quad
    \mathbf{z}\sim\mathcal{N}(\mathbf{0},\mathbf{I}),
\end{equation}
where $(\mathbf{x}_0,\mathbf{x}_1)\sim\pi$ has marginals $\rho_0,\rho_1$, $\mathbf{u}$ is an
auxiliary latent variable, $I$ satisfies the endpoint conditions, and typically $\gamma(0)=\gamma(1)=0$.
The induced density $\rho_t$ satisfies a transport equation and a family of forward/backward
Fokker--Planck equations with a tunable diffusion coefficient $\epsilon(t)$, leading to associated
ODE/SDE samplers (e.g., a forward SDE of the form $d\mathbf{x}_t=\mathbf{b}(\mathbf{x}_t,t)dt+\sqrt{2\epsilon(t)}\,d\mathbf{W}_t$).
For simplicity, we assume independent couplings and omit $\mathbf{u}$.
In robotics, Chen et al.~\cite{chen24bridger} leverage \textit{conditional} stochastic interpolants to bridge from informative (non-Gaussian)
source policies, improving few-step policy diffusion performance. 

\mypara{Inference-time Guidance.}
\del{Inference-time guidance methods, originally developed for image synthesis, including  classifier guidance~\cite{dhariwal2021diffusion} and energy-based guidance~\cite{graikos2022diffusion}, modify the sampling process using gradients derived from auxiliary \del{cost functions} \rev{costs or likelihoods}.} \rev{Inference-time guidance methods, popularized in image synthesis through classifier guidance~\cite{dhariwal2021diffusion} and related energy- or loss-based guidance schemes~\cite{graikos2022diffusion, 10377605}, modify the sampling process using gradients derived from auxiliary likelihoods, costs, or differentiable constraints. Related formulations include posterior-sampling methods with diffusion or flow priors, e.g., DPS~\cite{chung2023diffusion}, DAPS~\cite{zhang2025improving}, and FlowDPS~\cite{kim2025flowdps}, as well as optimal-control-based flow guidance such as OC-Flow~\cite{wang2025training}; these methods mainly target inverse problems or general controlled generation.}

In robotics, \del{such methods have been applied to} \rev{inference-time guidance has been used to} enforce safety constraints\rev{, temporal-logic specifications, } and kinematic requirements~\cite{wang2025inference, feng2024ltldog, yan2025m, deng2025safebimanual, hao2025disco}. Most existing approaches rely on diffusion- or flow-based policies~\cite{chi2023diffusion, feng2025guidance, janner2022planning} that generate complete action chunks in a single inference step. While effective for static objectives, this open-loop execution paradigm exhibits the \emph{commitment problem}~\cite{xue2025reactive}, limiting responsiveness to time-varying constraints and dynamic environments.

\mypara{Our contributions.}
We extend SI-based policies~\cite{chen24bridger} to a \textit{guided streaming} formulation, improving efficiency and reactivity. In doing so, we develop a general inference-time conditioning framework for \emph{general} stochastic interpolants. When instantiated for robot control, this framework enables online adaptation to test-time constraints. From a complementary perspective, we generalize SFP from deterministic to stochastic dynamics, which permits a principled derivation of inference-time guidance using the Feynman--Kac representation and the Doob \(h\)-transform.

\rev{This stochastic-path perspective provides a natural route to guided streaming control, where stochasticity serves as the mathematical structure enabling the guidance derivation rather than as a claim of superior performance over flow matching.}

\section{Guided Streaming Stochastic Interpolant Policies (Guided SSIP)}
\label{sec:method}

Our main objective is to sample from a conditional target distribution defined over complete trajectories $\target(\traj \mid \xi)$, where $\traj$ denotes a trajectory over the horizon $[0,T]$ (a path in action/state space) and $\xi$ is an exogenous context (e.g., a goal specification or an observation/history embedding). 
This target is constructed by reweighting a base dynamics distribution $\base(\traj \mid \xi)$ via a scalar cost function $\cost(\traj;\xi)$:
\begin{equation}
    \target(\traj \mid \xi) = \frac{1}{Z_\xi} \base(\traj \mid \xi) e^{-\cost(\traj;\xi)}
    \label{eq:global_target}
\end{equation}
where $Z_\xi = \int \base(\traj \mid \xi) e^{-\cost(\traj;\xi)} d\traj$ is the global partition function. The cost function $\cost(\traj; \xi)$ encapsulates any inference-time objective used to guide the robot, such as obstacle avoidance or alignment with human preferences. 

In this section, we establish the theoretical framework for Guided Streaming Stochastic Interpolant Policies.
First, we derive the optimal guidance law for \textit{general} stochastic interpolants (Section \ref{subsec:general_guidance}).
Next, we introduce our specific architecture, the Streaming Stochastic Interpolant Policy (SSIP), which generalizes the deterministic Streaming Flow Policy (SFP) into a Stochastic Differential Equation (SDE) to provide the necessary support for this guidance (Section \ref{subsec:ssip}).
Finally, we bridge global trajectory objectives and instantaneous control by deriving the action target conditional over the current time step (Section~\ref{subsec:marginal}). This yields a utility function that is compatible with the martingale-based guidance law, justifying its direct application to the streaming setting (Section~\ref{guide_streaming}).

\subsection{Optimal Guidance for Stochastic Interpolants}
\label{subsec:general_guidance}

We derive the exact guidance law for a general conditional Stochastic Interpolant \cite{albergo2023stochastic} whose base dynamics evolve over a generic state $\mathbf{x}_t$ and are governed by the diffusion
\begin{equation}
    d\mathbf{x}_t = \mathbf{b}(\mathbf{x}_t,t,\xi)\,dt + \sqrt{2\epsilon(t)}\,d\mathbf{W}_t,
    \label{eq:base_sde}
\end{equation}
where $\mathbf{b}$ is the base drift, $\epsilon(t)\ge 0$ is an isotropic diffusion schedule, and $\mathbf{W}_t$ is a standard Wiener process. Our global objective (Eq.~\eqref{eq:global_target}) defines a \emph{path-space} reweighting of the base trajectory measure conditional on $\xi$:
\begin{equation}
    p_{\text{target}}(\tau_{0:T}\mid \xi) \propto p_{\text{base}}(\tau_{0:T}\mid \xi)e^{-J(\tau_{0:T},\xi)},
    \label{eq:path_tilt}
\end{equation}
where $J(\tau_{0:T},\xi)$ is a trajectory-level cost. For the remainder of this subsection, we assume a standard \emph{time-additive} form
\begin{equation}
    J(\tau_{0:T},\xi) \;=\; \int_0^T c(\mathbf{x}_s,s,\xi)\,ds \;+\; \phi(\mathbf{x}_T,\xi),
    \label{eq:additive_cost}
\end{equation}
which includes pure terminal costs as the special case $c\equiv 0$.

\mypara{Desirability and the martingale condition.} Let $\{\mathcal{F}_t\}$ denote the natural filtration of the base process \eqref{eq:base_sde}. Define the \emph{global} likelihood-ratio process
\begin{align}
    D_t \triangleq\frac{\mathbb{E}_{\text{base}}\left[e^{-J(\tau_{0:T},\xi)}\mid \mathcal{F}_t\right]}{Z(\xi)}
    \label{eq:likelihood_ratio}
\end{align}
where $Z(\xi) \triangleq\mathbb{E}_{\text{base}}\left[\exp\left(-J(\tau_{0:T},\xi)\right)\right]$.
By construction and the tower property, $\{D_t\}_{t\in[0,T]}$ is a (nonnegative) martingale with $\mathbb{E}[D_t]=1$. Under the additivity assumption \eqref{eq:additive_cost} and Markovity of \eqref{eq:base_sde}, the conditional expectation in \eqref{eq:likelihood_ratio} factorizes into a \emph{sunk} past factor and a \emph{future} desirability:
\begin{align}
    \mathbb{E}_{\text{base}}&\left[e^{-J(\tau_{0:T},\xi)}\mid \mathcal{F}_t \right] = \nonumber \\
    & \quad\exp\left(-\int_0^t c(\mathbf{x}_s,s,\xi)ds\right)u(\mathbf{x}_t,t,\xi),
    \label{eq:martingale_factorization}
\end{align}
where the desirability (a.k.a. exponentiated future utility) is
\begin{align}
    &u(\mathbf{x},t,\xi)
    \triangleq \nonumber \\
    & \,\mathbb{E}_{\text{base}}\left[\exp\left(-\int_t^T c(\mathbf{x}_s,s,\xi)\,ds - \phi(\mathbf{x}_T,\xi)\right)\Bigm|\mathbf{x}_t=\mathbf{x}\right] 
    \label{eq:desirability_def}
\end{align}

Applying It\^{o}'s lemma to the martingale in \eqref{eq:martingale_factorization} yields the (backward) Feynman--Kac equation satisfied by $u$:
\begin{align}
    \partial_t u(\mathbf{x},t,\xi) + \mathcal{L}_t u(\mathbf{x},t,\xi) - c(\mathbf{x},t,\xi)\,u(\mathbf{x},t,\xi) & = 0,
    \label{eq:feynman_kac}
\end{align}
with $u(\mathbf{x},T,\xi)=e^{-\phi(\mathbf{x},\xi)}$, and the time-dependent generator
\begin{equation}
    \mathcal{L}_t f \;=\; \mathbf{b}(\mathbf{x},t,\xi)\cdot \nabla_{\mathbf{x}} f \;+\; \epsilon(t)\,\Delta_{\mathbf{x}} f.
    \label{eq:generator}
\end{equation}
In the pure terminal-cost case ($c\equiv 0$), \eqref{eq:feynman_kac} reduces to the homogeneous backward Kolmogorov equation $\partial_t u + \mathcal{L}_t u = 0$.

\mypara{Optimal drift via a Doob $h$-transform.}
We now characterize the dynamics under the tilted path measure \eqref{eq:path_tilt}. Using \eqref{eq:likelihood_ratio} as the Radon-Nikodym derivative on $\mathcal{F}_t$, It\^{o} calculus together with \eqref{eq:feynman_kac} gives
\begin{equation}
    dD_t \;=\; D_t\,\sqrt{2\epsilon(t)}\,\nabla_{\mathbf{x}} \log u(\mathbf{x}_t,t,\xi)\cdot d\mathbf{W}_t.
    \label{eq:D_martingale_sde}
\end{equation}
By Girsanov's theorem, under the tilted measure the process remains a diffusion with the \emph{same} diffusion coefficient and a shifted drift:
\begin{equation}
    d\mathbf{x}_t
    =
    \underbrace{\Bigl[\mathbf{b}(\mathbf{x}_t,t,\xi) + 2\epsilon(t)\,\nabla_{\mathbf{x}}\log u(\mathbf{x}_t,t,\xi)\Bigr]}_{\tilde{\mathbf{b}}(\mathbf{x}_t,t,\xi)}\,dt
    +
    \sqrt{2\epsilon(t)}\,d\widetilde{\mathbf{W}}_t, \nonumber
    \label{eq:guided_sde}
\end{equation}
where $\widetilde{\mathbf{W}}_t$ is a Wiener process under the tilted measure. Therefore, the exact guidance modification is
\begin{align}
    \Delta \mathbf{b}(\mathbf{x},t,\xi)
    \triangleq
    \tilde{\mathbf{b}}(\mathbf{x},t,\xi)-\mathbf{b}(\mathbf{x},t,\xi)
    =
    2\epsilon(t)\,\nabla_{\mathbf{x}}\log u(\mathbf{x},t,\xi).
    \label{eq:optimal_guidance}
\end{align}
This result is the continuous-time analogue of KL-minimal control; we obtain the target path reweighting \eqref{eq:path_tilt} by adding the smallest possible (in relative-entropy) drift correction consistent with the tilted measure.

\subsection{Streaming Stochastic Interpolant Policy (SSIP)}
\label{subsec:ssip}
We define SSIP as an execution-time-aligned {diffusion policy} whose time-indexed marginals follow a {stochastic interpolant} path. This is strictly more general than any particular deterministic streaming flow; deterministic policies arise as special cases (e.g., drift-only samplers or zero-noise interpolants), while the diffusion formulation is what enables principled guidance via the stochastic control results (in Section~\ref{subsec:general_guidance}).

Here, the diffusion state is $\mathbf{x}_t=\mathbf{a}_t$ and context $\xi=\mathbf{h}$ denotes a snapshot of the policy context at the current control cycle (e.g., observation/history embedding). Within a single SSIP rollout over $t\in[0,T]$, we treat $\mathbf{h}$ as fixed, similar to SFP~\cite{jiang2025streaming}. Within each action chunk, while the history $\mathbf{h}$ remains fixed, each new drift is computed using the last action as input. This ensures reactivity, unlike conventional methods where the entire chunk is computed simultaneously.

Then, a stochastic interpolant policy specifies a family of conditional marginals $\{p_t(\mathbf{a}\mid \mathbf{h})\}_{t\in[0,T]}$ by constructing a noised sample as
\begin{equation}
    \mathbf{a}_t = \mathcal{I}_t(\mathbf{h},\zeta) + \gamma(t)\,\mathbf{z},
    \qquad
    \mathbf{z}\sim\mathcal{N}(\mathbf{0},\mathbf{I}),
    \label{eq:ssip_interpolant}
\end{equation}
where:
(i) $\mathcal{I}_t(\mathbf{h},\zeta)$ is a \emph{reference interpolant} (``signal path'') that may be deterministic given $\mathbf{h}$ or may depend on a latent $\zeta$ capturing structured trajectory variation, and
(ii) $\gamma(t)$ is a smooth scalar noise schedule controlling the marginal perturbation scale (often chosen to be small near the endpoints and larger mid-horizon).
Note that Eq.~\eqref{eq:ssip_interpolant} defines a \emph{path of marginals} used for training (by sampling fresh $\mathbf{z}$ for each $(t,\mathbf{h})$); it does not mean that a single fixed $\mathbf{z}$ is carried through time as a Brownian path. The continuous-time sampler is an SDE driven by a Wiener process $d\mathbf{W}_t$ (defined below).  \rev{Eq.~(13) is the policy-level instance of Eq.~(1): the generic state $\mathbf{x}_t$ corresponds to the action $\mathbf{a}_t$, and the interpolant is conditioned on the policy context $\mathbf{h}$.}

To realize the marginal path induced by \eqref{eq:ssip_interpolant} at inference time, we use a diffusion with drift decomposed into a transport term and a score term. Let
\[
s_\theta(\mathbf{a},t,\mathbf{h}) \approx \nabla_{\mathbf{a}}\log p_t(\mathbf{a}\mid \mathbf{h})
\]
denote a learned score, and let $v_\theta(\mathbf{a},t,\mathbf{h})$ denote a learned transport/velocity field.
We define the SSIP sampling dynamics as the one-parameter family
\begin{align}
    d\mathbf{a}_t = \Big[
        v_\theta(\mathbf{a}_t,t,\mathbf{h})
        +  \bigl(\epsilon(t)-&\gamma(t)\dot{\gamma}(t)\bigr)\,s_\theta(\mathbf{a}_t,t,\mathbf{h})
    \Big]dt + \nonumber \\
     & \sqrt{2\epsilon(t)}\,d\mathbf{W}_t,
    \label{eq:ssip_sde}
\end{align}
where $\epsilon(t)\ge 0$ controls the \emph{amount of stochasticity} injected during sampling.
The term $-\gamma(t)\dot{\gamma}(t)\,s_\theta$ is the deterministic  correction associated with the marginal noising schedule $\gamma(t)$; the additional $\epsilon(t)\,s_\theta$ and $\sqrt{2\epsilon(t)}\,d\mathbf{W}_t$ provide a tunable stochastic sampler. In the idealized setting where $s_\theta$ equals the true score, different choices of $\epsilon(t)$ correspond to different samplers that target the same marginal path, trading off exploration versus determinism.

SSIP cleanly separates two conceptually different kinds of ``randomness'':
\begin{itemize}
    \item {Structured variation (model diversity):} randomness in $\mathcal{I}_t(\mathbf{h},\zeta)$ (e.g., latent-conditioned behaviors or multimodal skills). This represents genuine alternative behaviors the policy has learned.
    \item {Sampling diffusion (exploration noise):} stochasticity controlled by $\epsilon(t)$ and realized by the Wiener process $d\mathbf{W}_t$. This is an inference-time knob that enables exploration and supports principled guidance.
\end{itemize}
We can recover the deterministic streaming flow policy by using a drift-only sampler,
\begin{equation}
    \frac{d\mathbf{a}_t}{dt}
    =
    v_\theta(\mathbf{a}_t,t,\mathbf{h})
    - \gamma(t)\dot{\gamma}(t)\,s_\theta(\mathbf{a}_t,t,\mathbf{h}).
    \label{eq:ssip_ode}
\end{equation}
and choosing a noise-free interpolant ($\gamma(t)\equiv 0$). 
This is the deterministic counterpart of \eqref{eq:ssip_sde} and is the default at execution when low-noise trajectories are preferred.
Then, \eqref{eq:ssip_ode} reduces to $\dot{\mathbf{a}}_t = v_\theta(\mathbf{a}_t,t,\mathbf{h})$. Specific constructions such as ``Gaussian-tube'' stabilizing dynamics correspond to particular choices of $\mathcal{I}_t$ and $v_\theta$ (and optionally a latent $\zeta$), but are not required by the SSIP framework. We provide an extended SSIP formulation and training details in Appendix~\ref{app:ssip}.

\subsection{Derivation of the Target Conditional}
\label{subsec:marginal}

SSIP operates by sampling the current action/state at time $t$ from an implicit generative distribution. To apply the trajectory-level target in Eq.~\eqref{eq:global_target}, we derive the induced \emph{target conditional} over the current action given the realized past. For clarity, we write trajectories as $\tau_{0:T} = (\mathbf{a}_0,\dots,\mathbf{a}_T)$. At decision time $t$, the realized history up to but \emph{excluding} the current action is fixed, i.e.,
$\tau_{0:t-1}$ is given and $\mathbf{a}_t$\ is the decision variable\footnote{We will work with discrete time-steps, but the derivation for continuous time is identical if we replace $\tau_{0:t-1}$ with the filtration $\mathcal{F}_t$.}.

As before, the global target over complete trajectories is
\begin{equation}
    p_{\text{target}}(\tau_{0:T}\mid\xi) =
    \frac{1}{Z_\xi}\,p_{\text{base}}(\tau_{0:T}\mid\xi)e^{-J(\tau_{0:T},\xi)}.
    \label{eq:target_path}
\end{equation}
For readability, we suppress $\xi$ in the remainder of the derivation in this subsection.
By marginalizing future trajectories $\tau_{t+1:T}$, the target conditional over the current action is
\begin{equation}
    p_{\text{target}}(\mathbf{a}_t \mid \tau_{0:t-1}) =
    \int p_{\text{target}}(\tau_{t:T}\mid \tau_{0:t-1})\,d\tau_{t+1:T}.
    \label{eq:target_action_marginal_def}
\end{equation}
Using \eqref{eq:target_path} and dropping normalization constants that do not depend on $\mathbf{a}_t$ yields
\begin{align}
    p_{\text{target}}& (\mathbf{a}_t \mid \tau_{0:t-1})
    \propto \nonumber \\
    & \int p_{\text{base}}(\tau_{t:T}\mid \tau_{0:t-1})e^{-J(\tau_{0:T})}d\tau_{t+1:T}.
    \label{eq:target_action_marginal_un}
\end{align}
We can factor the base conditional into two components,
\begin{align}
    p_{\text{base}}&(\tau_{t:T}\mid \tau_{0:t-1}) = \nonumber \\
    & \underbrace{p_{\text{base}}(\mathbf{a}_t\mid \tau_{0:t-1})}_{\text{Current Action}}\,
    \underbrace{p_{\text{base}}(\tau_{t+1:T}\mid \tau_{0:t-1}, \mathbf{a}_t)}_{\text{Future Rollout}}.
    \label{eq:base_factor}
\end{align}
Substituting \eqref{eq:base_factor} into \eqref{eq:target_action_marginal_un} and pulling out the $\mathbf{a}_t$-dependent term gives
\begin{align}
    p&_{\text{target}}(\mathbf{a}_t \mid \tau_{0:t-1})
    \propto \nonumber\\
    & p_{\text{base}}(\mathbf{a}_t\mid \tau_{0:t-1})
    \mathbb{E}_{\tau_{t+1:T}\sim p_{\text{base}}(\cdot\mid \tau_{0:t-1},\mathbf{a}_t)}
    \left[e^{-J(\tau_{0:T})}\right].
    \label{eq:target_action_marginal_expectation}
\end{align}
Under additive costs (Eq.~\eqref{eq:additive_cost}), the trajectory cost splits as
$J(\tau_{0:T}) = J(\tau_{0:t-1}) + J(\tau_{t:T})$,
where $J(\tau_{0:t-1})$ is fully determined by the fixed past. Therefore,
\begin{align}
\mathbb{E}\Big[e^{-J(\tau_{0:T})} & \mid \tau_{0:t-1},\mathbf{a}_t\Big] \nonumber \\
& =
e^{-J(\tau_{0:t-1})}
\mathbb{E}\left[e^{-J(\tau_{t:T})}\mid \tau_{0:t-1},\mathbf{a}_t\right],
\end{align}
and the prefactor $e^{-J(\tau_{0:t-1})}$ cancels in \eqref{eq:target_action_marginal_expectation} because it is independent of $\mathbf{a}_t$.
Define the (unnormalized) \emph{utility} of the current action as the expected exponentiated \emph{future} cost:
\begin{equation}
    u(\mathbf{a}_t, t, \tau_{0:t-1})
    \triangleq
    \mathbb{E}_{\tau_{t+1:T}\sim p_{\text{base}}(\cdot\mid \tau_{0:t-1},\mathbf{a}_t)}
    \left[e^{-J(\tau_{t:T})}\right]
    \label{eq:utility_def_streaming}
\end{equation}
Then the exact target conditional is
\begin{equation}
    p_{\text{target}}(\mathbf{a}_t \mid \tau_{0:t-1})
    \;\propto\;
    p_{\text{base}}(\mathbf{a}_t\mid \tau_{0:t-1})\;u(\mathbf{a}_t,t,\tau_{0:t-1}).
    \label{eq:target_action_final}
\end{equation}

\subsection{Guiding Streaming Policies}
\label{guide_streaming}

To guide SSIP, we apply the general result of Section~\ref{subsec:general_guidance} to the SSIP base sampler \eqref{eq:ssip_sde}. Define the SSIP base drift
\begin{equation}
\mathbf{b}_\theta(\mathbf{a},t,\mathbf{h})
\triangleq
v_\theta(\mathbf{a},t,\mathbf{h})
+\bigl(\epsilon(t)-\gamma(t)\dot{\gamma}(t)\bigr)s_\theta(\mathbf{a},t,\mathbf{h}).
\label{eq:ssip_base_drift}
\end{equation}
Let the streaming desirability (future utility) be
\begin{equation}
u(\mathbf{a}_t,t,\mathbf{h})
\triangleq
\mathbb{E}_{\tau_{t+1:T}\sim p_{\text{base}}(\cdot\mid \mathbf{a}_t,t,\mathbf{h})}
\!\left[\exp\!\left(-J(\tau_{t:T})\right)\right].
\label{eq:streaming_u}
\end{equation}
Then the guided SSIP dynamics are obtained by adding the optimal correction $\Delta \mathbf{b}=2\epsilon(t)\nabla_{\mathbf{a}}\log u$:
\begin{align}
d\mathbf{a}_t
=
\Bigl[
\mathbf{b}_\theta(\mathbf{a}_t,t,\mathbf{h})
+ 2\epsilon(t)\nabla_{\mathbf{a}}\log u(\mathbf{a}_t,&t,\mathbf{h})
\Bigr]dt + \nonumber \\
& \sqrt{2\epsilon(t)}\,d\mathbf{W}_t.
\label{eq:guided_ssip_final}
\end{align}
Reactivity is ensured by receding-horizon execution: after executing an action from the current rollout, we compute the next step based on the current action $\mathbf{a}_t$ and the utility calculated at that specific timestep, which reflects the instantaneous future trajectory cost $J(\tau_{t:T})$.
In practice, the remaining challenge is estimating $\nabla_{\mathbf{a}}\log u$ efficiently at inference time, which we address in the following section.

\section{Guidance for On-the-fly Obstacles}
\label{sec:guidance_obstacles}

As shown above, the optimal guidance drift is proportional to the gradient of the log-expected future utility: $\nabla_{\mathbf{a}} \log \mathbb{E}[e^{-J(\tau_{t:T})}]$. While this formulation provides a theoretical foundation for constraint satisfaction, computing this gradient exactly is often computationally intractable, as it necessitates differentiating through an expectation over a high-dimensional distribution of future trajectories.
To bridge this gap, we propose two complementary strategies tailored to different deployment requirements, described below.

\subsection{Stochastic Trajectory Ensemble Guidance (STEG)}
\label{sec:teg}

To enable zero-shot adaptation, we propose Stochastic Trajectory Ensemble Guidance (STEG). This method leverages the differentiability of our SSIP backbone to construct a Monte Carlo estimator of the utility gradient $\nabla_{\mathbf{a}} \log \mathbb{E}[e^{-J(\tau)}]$ on-the-fly. Estimating the guidance gradient using a finite set of sample trajectories introduces two primary sources of error: \textit{Distributional Shift} (failure to cover safe modes) and \textit{Dynamics Mismatch} (incorrect gradient direction). In {Appendix \ref{app:estimator_theory}}, we provide an error decomposition of the guidance score. Our analysis suggests two critical design principles for a valid estimator:
\begin{itemize}
    \item {High-Fidelity Support ($N \uparrow$):} The estimator must employ a large ensemble size to minimize the selection error arising from the multi-modal nature of the safety posterior.
    \item {Physics-Consistent Gradients:} The rollout mechanism must preserve causal derivatives ($\partial \mathbf{a}_{t+k} / \partial \mathbf{a}_t$) to ensure the guidance force respects system kinematics.
\end{itemize}
STEG implements these principles by combining parallel sampling with differentiable physics.

At inference time, STEG approximates the intractable expectation $\mathbb{E}_{\tau}[e^{-J(\tau)}]$ by simulating a batch of $N$ parallel chains. We use the learned SDE dynamics itself as the transition model, conditioning on the snapshot context $\mathbf{h}$ and holding it fixed throughout the $K$-step rollout. The process consists of three steps:
\begin{enumerate}
\item {Stochastic Ensemble Rollout.} We replicate the current state $\mathbf{a}_t$ into an ensemble $\{\mathbf{a}_t^{(i)}\}_{i=1}^N$. We then unroll the dynamics for $K$ steps using the Euler-Maruyama discretization of the base SSIP generative SDE.

\item {Utility Aggregation.} During the rollout, we accumulate the running cost for each trajectory $J(\tau^{(i)}) = \sum_{k=1}^K c(\mathbf{a}_k^{(i)}) \Delta t$, where $c(\cdot)$ is the additive cost function (e.g., repulsive potential). We then estimate the log-expected utility by employing the {Log-Sum-Exp (LSE)} estimator:
\begin{align}
    \hat{\mathcal{V}}(\mathbf{a}_t) & = \text{LogSumExp}\left( \{-J(\tau^{(i)})\}_{i=1}^N \right) - \log N \nonumber \\
    & \approx \log \mathbb{E}[e^{-J}]
\end{align}
\item {Backpropagation Through Time (BPTT).}
The optimal guidance drift is the gradient of this value estimate. Leveraging automatic differentiation, we compute the gradient w.r.t. the current action $\mathbf{a}_t$ by backpropagating through the unrolled graph:
\begin{equation}
    \Delta \mathbf{b}(\mathbf{a}_t, t, \mathbf{h}) = 2\epsilon(t) \cdot \nabla_{\mathbf{a}_t} \hat{\mathcal{V}}(\mathbf{a}_t)
\end{equation}
\end{enumerate}

In summary, STEG provides a principled, training-free mechanism for inference-time guidance by directly estimating the utility gradient through differentiable stochastic rollouts. STEG enables robust zero-shot adaptation to novel objectives while preserving the smoothness and reactivity of the underlying SSIP dynamics. This makes STEG well-suited for unstructured and rapidly changing environments, where safety constraints and task objectives cannot be anticipated at training time.

\subsection{Conditional Critic Guidance (CCG)}
\label{subsec:ccg}

To achieve real-time reactivity, we propose {Conditional Critic Guidance (CCG)}, where we train a parameterized critic network $V_\psi$ to approximate the guidance potential directly.
Since the base SSIP policy generates actions autoregressively, the distribution of future trajectories is inherently conditioned on the robot's history. Consequently, the value function must account for this temporal context. We parameterize the critic as $V_\psi(\mathbf{a}_t, t, \mathbf{h}, \phi)$, where $\mathbf{h}$ represents the history context (e.g., observation history) used by the base policy, and $\phi$ represents the parameters defining the cost (e.g., human preferences or obstacle positions). In the following sections, we specifically use $\phi$ to denote obstacle parameters. The network is trained to approximate a scalar value $y_\text{target}$ derived from future rollouts:
\begin{equation}
    V_\psi(\mathbf{a}_t, t, \mathbf{h}, \phi) \approx y_\text{target}(\mathbf{a}_t, \mathbf{h}, \phi)
\end{equation}

\rev{Here, $y_{\text{target}}$ estimates the log-desirability $\log u(\mathbf{a}_t,t,\mathbf{h})$ in Eq.~\eqref{eq:streaming_u}, whose gradient yields the guidance correction in Eq.~\eqref{eq:guided_ssip_final} for inducing the target conditional in Eq.~\eqref{eq:target_action_final}.}

In practice, CCG is most effective in structured environments where obstacle classes and cost functions are known at training time. By amortizing the guidance computation into a learned critic, CCG enables low-latency inference when prior knowledge is available.

\rev{
\subsection{Summary of the Main Method}

In summary, Guided SSIP proceeds in three steps. First, we instantiate the streaming SI sampler with the base drift in Eq.~\eqref{eq:ssip_base_drift}. Second, at each execution step, we estimate the log-desirability $\log u(\mathbf{a}_t,t,\mathbf{h})$: STEG estimates it online through short-horizon stochastic rollouts, while CCG amortizes it with a trained critic. Third, we apply the guidance correction in Eq.~\eqref{eq:guided_ssip_final} and execute the resulting action in a receding-horizon manner. The full inference procedures are given in Alg.~\ref{alg:steg_inference} for STEG and Alg.~\ref{alg:ccg_inference} for CCG.
}

\section{Experiments}
\label{sec:experiments}

We designed our experiments to investigate the following research questions:
\begin{enumerate}
    \item[(\textbf{Q1})] How does our method compare to conventional guidance techniques in static environments?
    \item[(\textbf{Q2})] Does our method demonstrate adaptability when encountering dynamic or unexpectedly appearing obstacles compared to baseline techniques?
    \item[(\textbf{Q3})] What is the computational overhead of STEG compared to CCG with pre-trained critics, and what are the respective strengths and weaknesses of each approach in real-time scenarios?
\end{enumerate}

\subsection{Experimental Setup}
To answer these questions, we integrate our streaming guidance techniques with the SSIP and evaluate them on obstacle avoidance in two \del{different} \rev{simulation} domains: the {Push-T} task and simulated robot manipulation in {Robosuite}. All base policies are trained on datasets containing \textit{only} \textit{collision-free} demonstrations. During evaluation, we introduce unmodeled static and dynamic obstacles to assess the  capabilities of our proposed guidance framework. 

\mypara{Baselines and SSIP Variants.} We compare our methods against two baselines: conventional action chunk-based diffusion policy ({DP}) with energy-based guidance~\cite{janner2022planning} and flow-policy ({FP}) with flow-guidance~\cite{feng2025guidance}. For SSIP, we compare the STEG and CCG guidance methods against a simple repulsion method (a simple additive residual control based on the distance to obstacles that works well in practice).
\del{For STEG guidance, we limit rollouts to $K=3$ steps for Push-T and $K=4$ steps for Robomimic.} \rev{For STEG guidance, we use $K=3$ rollout steps with $N=64$ parallel rollouts for Push-T and $K=4$ rollout steps with $N=32$ parallel rollouts for Robomimic.} For CCG guidance, we employ two different critics trained with the following costs:
\begin{enumerate}
    \item {Distance Potential Cost (CCG-D):} the regression target approximates the log-expected future utility $\log \mathbb{E}[e^{-J}]$ by aggregating cumulative distance costs from rollouts. 
    \item {Collision Proxy Cost (CCG-P):} the target is a collision-risk proxy computed from future rollouts (based on \textit{minimum} distance and approach angle). This relaxes the additive-cost assumption required for the exact Doob-transform guidance, but provides a practical, reactive guidance field.
\end{enumerate}

\mypara{Evaluation Protocols.}
Guided policies exhibit varying sensitivity to the guidance scale \(\lambda\), which controls the trade-off between task performance and safety. To ensure a fair and comprehensive comparison, we adopt two complementary evaluation protocols:
\begin{itemize}
\item {Peak Performance Metrics.}
We perform a grid search over \(\lambda\) and report results at the configuration that achieves the highest task success rate (SR). A trajectory is considered successful if its final reward exceeds a task-specific threshold (85\%, corresponding to a slight relaxation to account for necessary avoidance maneuvers). We additionally report inference latency, defined as the average wall-clock time required to generate a single action step\footnote{All evaluations were conducted on a workstation with an AMD EPYC 7543 processor and a single A5000 GPU.}. 
\item {Safety-Reward Trade-off.}
We sweep \(\lambda\) and report the Pareto frontier of average reward versus safety rate (\(1 - \text{Collision Rate}\)). Methods closer to the top-right corner exhibit stronger task performance while maintaining higher safety.
\end{itemize}

        
        
        


\begin{table}[t]
    \centering
    \caption{Experimental results on the Push-T task with static obstacles. We compare Task Success Rate (SR) and Inference Latency. Our guided SSIP methods outperform the baselines in terms of success rate with lower latency. \textbf{Bold} indicates the best performance.}
    \label{tab:pusht_results}
    \small 
    \setlength{\tabcolsep}{4pt} 
    \begin{tabular}{ll cc}
        \toprule
        \textbf{Category} & \rev{\textbf{Method}} & \textbf{SR\%} ($\uparrow$) & \textbf{Latency} (ms) \\
        \midrule
        
        \multirow{2}{*}{Baseline} 
        & DP     & 58.5 & 91.52 \\
        & FP      & 56.0 & \textbf{7.78} \\

        \midrule
        
        \multirow{4}{*}{SSIP} 
        & Repulsion   & 68.6 & 11.37 \\
        & STEG           & 66.7 & 18.49 \\
        & CCG-D            & 65.7 & 13.72 \\
        & CCG-P     & \textbf{74.3} & 14.75 \\
        \bottomrule
    \end{tabular}
\end{table}

        

        
        
        
        

\begin{figure}[t]
    \centering
    \includegraphics[width=0.40\textwidth]{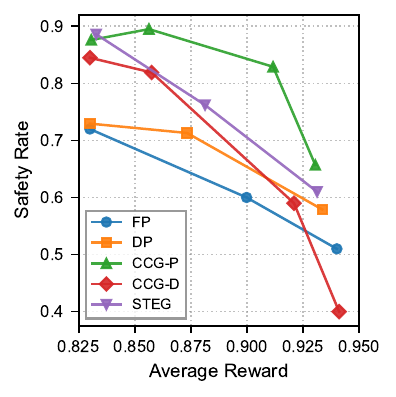}
    \caption{Safety-Reward trade-off. SSIP consistently achieves more favorable trade-offs compared to the baselines, with its Pareto frontiers closer to the top-right corner of the reward–safety plane. Among the SSIP variants, CCG-P dominates across the evaluated range, while STEG outperforms the CCG-D variant.}
    \label{fig:tradeoff1}
\end{figure}

\begin{figure}
    \centering
    \includegraphics[width=0.48\textwidth]{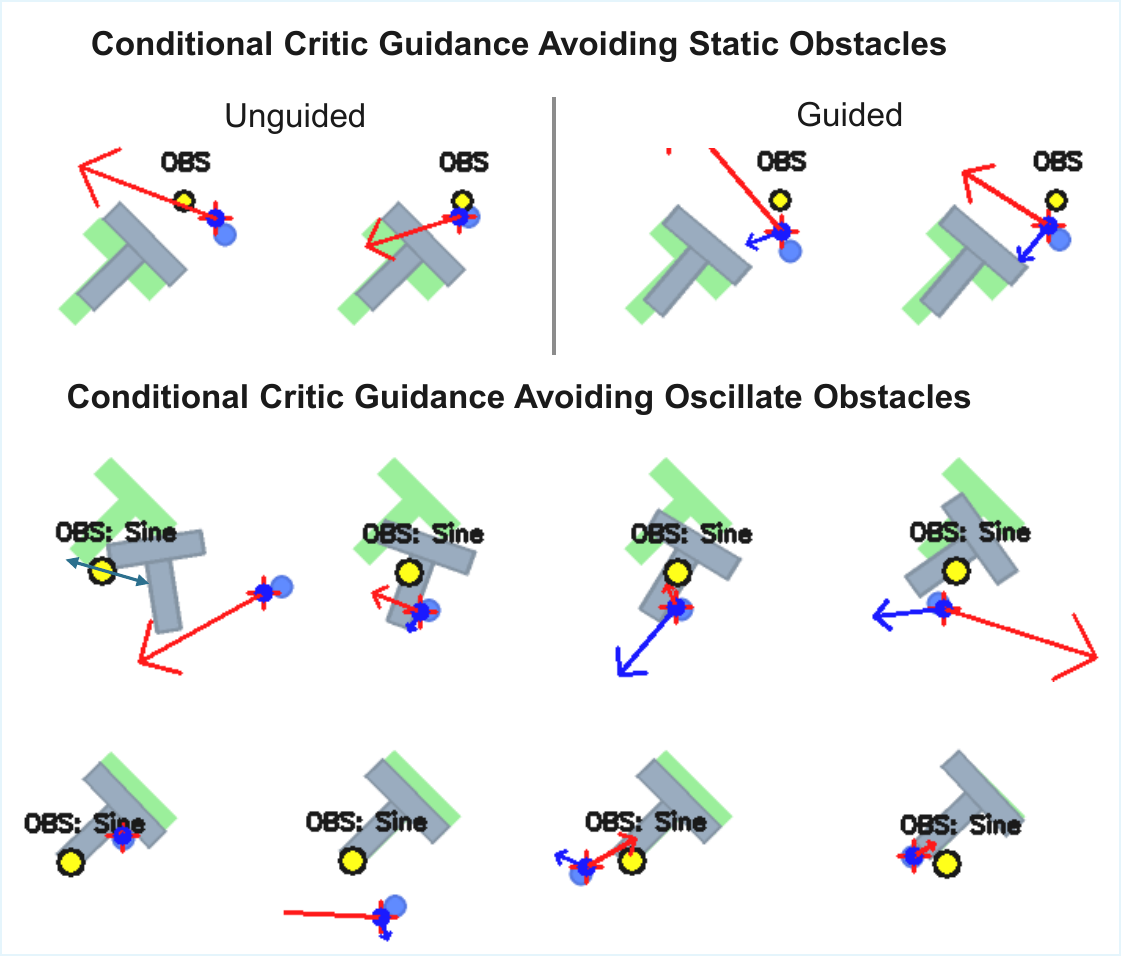}
    \caption{SSIP with CCG-P guidance in static and dynamic obstacle tasks. Red arrows ($\textcolor{red}{\rightarrow}$) represent the base policy while blue arrows ($\textcolor{blue}{\rightarrow}$) represent guidance. We can observe the guidance acts to push the end-effector away from the obstacles.}
    \label{fig:visual}
\end{figure}

\subsection{Push-T with Static Obstacles}
On this task, the robot has to complete the task (push the T-block into a particular goal configuration) without colliding with any of eight static obstacles. 
We execute the base policy and filter for instances where it achieves task success but experiences collisions, yielding a set of approximately 100 cases per method. 

\mypara{Results.}
Table~\ref{tab:pusht_results} reports success rate and inference latency on Push-T with static obstacles. Overall, SSIP improves task success relative to conventional chunk-based guidance baselines, addressing \textbf{Q1}. In particular, CCG-P achieves the highest success rate (74.3\%) while maintaining low per-step latency (14.75\,ms), outperforming classifier-guided DP (58.5\%, 91.52\,ms) and flow-guided FP (56.0\%, 7.78\,ms). Notably, a simple repulsion residual already performs strongly in this static setting (68.6\%, 11.37\,ms), indicating that local distance-based avoidance is often sufficient when obstacles are fixed.

Crucially, the per-step latency reported for conventional baselines represents an amortized metric. To reflect the true control latency, this value must be multiplied by the execution chunk size ($H_{execute}{=}8$), as these methods generate the entire action sequence simultaneously. Consequently, the robot faces a ``wait-then-act'' bottleneck resulting in unsmooth execution, a limitation also identified in SFP~\cite{jiang2025streaming}.

The distance-based instantiations of our principled guidance---CCG-D (65.7\%, 13.72\,ms) and STEG with short-horizon rollouts \(K{=}3\) (66.7\%, 18.49\,ms)---are comparable to, and slightly below, repulsion: they optimize essentially the same distance signal but incur additional approximation error from critic regression (CCG-D) or finite-sample, short-horizon Monte Carlo estimation (STEG). This highlights a key trade-off for \textbf{Q3}: CCG amortizes guidance into a critic for lower-latency inference, whereas STEG adds overhead due to online rollouts but remains training-free and broadly applicable. 

The trade-off curves shown in Fig.~\ref{fig:tradeoff1} are consistent with this view. SSIP yields a more favorable safety--reward trade-off than the chunk-based baselines, with Pareto frontiers consistently closer to the top-right region of the reward--safety plane. Among SSIP variants, CCG-P dominates across the swept guidance scales, maintaining high safety at comparable (or higher) reward. STEG improves over conventional baselines and generally exceeds CCG-D, but exhibits a less favorable trade-off than CCG-P in the high-reward regime, indicating a sharper safety degradation as reward is pushed upward.

\subsection{Push-T with Dynamic Obstacles}
To test reactivity, we introduce dynamic obstacles during the robot's execution. \del{The obstacles follow three distinct movement patterns, increasing in difficulty: (i) moving towards and intercepting the trajectory midpoint, (ii) oscillating around the midpoint, and (iii) chasing the end effector.} \rev{The obstacles follow three distinct movement patterns, increasing in kinematic difficulty: \textit{Intercept} moves predictably toward the trajectory midpoint, \textit{Oscillate} moves back and forth around the robot's nominal path, and \textit{Chase} continuously follows the end-effector, making within-horizon adaptation most challenging.}

\mypara{Results.} 
Table~\ref{tab:pusht_dynamic} shows that dynamic obstacles substantially amplify the limitations of chunk-based baselines, highlighting the benefit of streaming reactivity. In the simplest \emph{Intercept} setting, DP and FP remain competitive (72--76\%), and SSIP with CCG-P matches the best baseline (76\%), suggesting that when obstacle motion is predictable and brief, chunk-based guidance can still recover. 

As obstacle dynamics become more adversarial, SSIP variants separate more clearly from the baselines. Under \emph{Oscillate}, CCG-P achieves the highest success (68\%), outperforming both DP (38\%) and FP (50\%), indicating that continuous, online guidance is particularly effective when constraints change within an execution horizon. The most challenging \emph{Chase} regime further stresses within-horizon adaptation; both chunk-based baselines collapse (12\% for DP, 8\% for FP), while SSIP methods retain markedly higher success (30--36\%). Notably, STEG performs best in \emph{Chase} (36\%), consistent with the advantage of training-free, on-the-fly gradient estimation when obstacle behavior deviates from training-time conditions.

Overall, these results suggest a regime shift; chunk-based guidance is brittle under rapidly changing constraints, while SSIP enables substantially higher success. CCG-P excels in moderately dynamic settings and STEG provides the strongest robustness in the most adversarial motion pattern.

\begin{table}[t]
    \centering
    \caption{Experimental results on the Push-T task with dynamic obstacles. We evaluate the Task Success Rate (SR) across three distinct obstacle motion patterns: moving towards the trajectory midpoint (Intercept), oscillating around the midpoint (Oscillate), and chasing the end-effector (Chase).}
    \label{tab:pusht_dynamic}
    \small 
    \setlength{\tabcolsep}{3pt} 
    \begin{tabular}{ll ccc}
        \toprule
        & & \multicolumn{3}{c}{\textbf{Success Rate (SR\%)} $\uparrow$} \\
        \cmidrule(lr){3-5}
        \textbf{Category} & \rev{\textbf{Method}} & \textbf{Intercept} & \textbf{Oscillate} & \textbf{Chase} \\
        \midrule
        
        \multirow{2}{*}{Baseline} 
        & DP      & 72.0 & 38.0 & 12.0 \\
        & FP      & \textbf{76.0} & 50.0 & 8.0 \\
        \midrule
        
        \multirow{3}{*}{SSIP} 
        & Repulsion   & 64.0 & 54.0 & 30.0 \\
        & CCG-P    & \textbf{76.0} & \textbf{68.0} & 30.0 \\
        & STEG           & 68.0 & 56.0 & \textbf{36.0} \\

        \bottomrule
    \end{tabular}
\end{table}

\subsection{Simulation Robot Experiment}
\label{sec:sim_experiment}

We conduct experiments on three canonical Robomimic tasks: \textit{Lift}, \textit{Can}, and \textit{Square}. We introduce unmodeled static obstacles that intersect the nominal demonstration trajectories, which requires the agent to deviate to succeed. We evaluate 100 trials per baseline per task.

\begin{table}
    \centering
    \caption{\textbf{Robomimic Simulation Results (Success Rate).} We report the Success Rate (SR \%) across three tasks. }
    \label{tab:robomimic_results}

    \small 
    \begin{tabular}{llccc}
        \toprule
        & & \multicolumn{3}{c}{\textbf{Success Rate (SR\%)} $\uparrow$} \\
        \cmidrule(lr){3-5}
        \textbf{Category} & \textbf{Method} & \textbf{Lift} & \textbf{Square} & \textbf{Can} \\
        \midrule
        
        \multirow{1}{*}{Baseline} 
        & DP & 71.0 & 58.0 & 51.0 \\
        & FP & 72.0 & 48.0 & 51.0 \\
        \midrule
        
        \multirow{3}{*}{SSIP} 
        & Repulsion & 35.0 & 32.0 & 14.0 \\
        & STEG & 45.0 & 50.0 & 51.0 \\
        & CCG-P & \textbf{73.0} & \textbf{79.0} & \textbf{78.0} \\
        \bottomrule
    \end{tabular}
\end{table}

\begin{figure}
    \centering
    \includegraphics[width=0.48\textwidth]{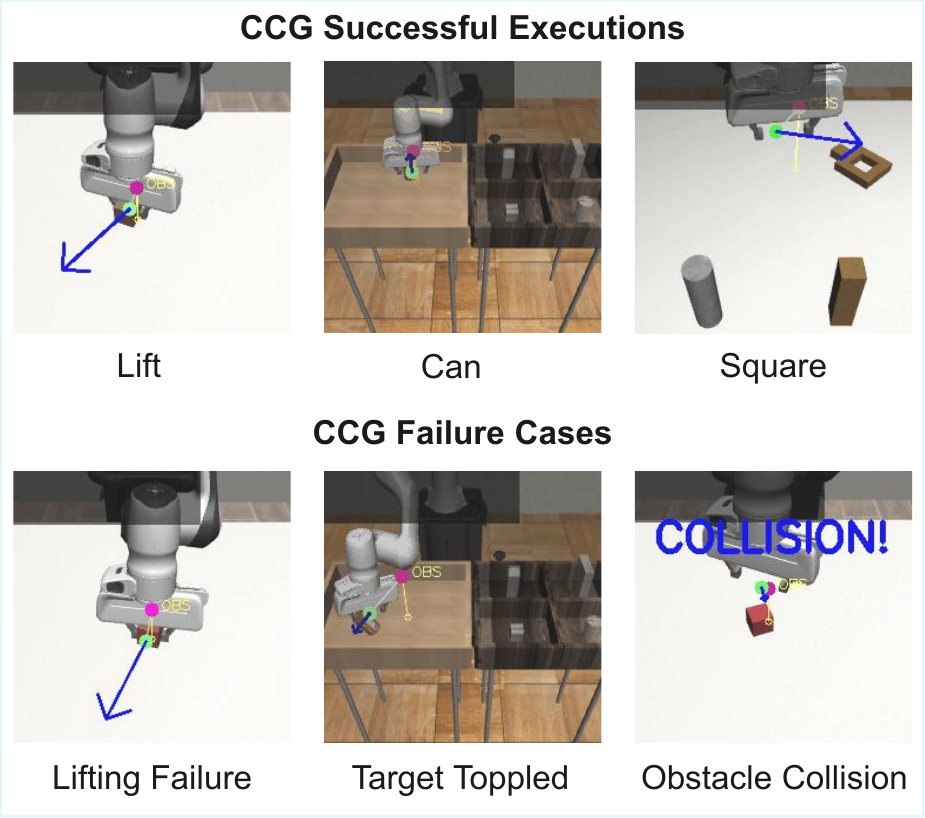}
    \caption{Visualization of CCG execution across three Robomimic tasks, including representative failure examples. Blue arrows indicate the guidance force.}
    \label{fig:robot}
\end{figure}

\mypara{Results.} Table~\ref{tab:robomimic_results} reports success rates across the three Robomimic tasks. Overall, SSIP with CCG-P guidance achieves the strongest performance, substantially outperforming all baselines across tasks. This result highlights the benefit of amortized, history-aware guidance when reliable prior structure is available.
Among the remaining SSIP variants, STEG consistently improves over the simple repulsion baseline but underperforms CCG-P. This gap is most pronounced in tasks where the obstacle directly intersects the nominal trajectory, suggesting that finite-horizon stochastic rollouts can struggle to discover sufficiently long-horizon detours under coarse discretization. Nevertheless, STEG still provides meaningful gains over heuristic guidance without requiring offline training.
In contrast, the repulsion-based method performs poorly across all tasks, indicating that local reactive penalties are insufficient to resolve obstacle interference while preserving task completion. This contrasts with the Push-T case, where simple repulsion did not significantly disrupt policy execution. Finally, conventional Diffusion and Flow Policies exhibit performance  better than STEG, indicating that they can still provide reasonable guidance forces in high-dimensional spaces when the reward function remains static within the chunk.

Analysis of failed episodes for CCG-P and STEG reveals two primary failure modes: \textit{insufficient avoidance} (leading to collision) and \textit{task interference} (leading to non-completion).
Collisions (Fig.~\ref{fig:robot}, bottom right) typically stem from the distributional constraints of the base policy. Since our guidance method operates by reweighting the learned trajectory distribution, it struggles to synthesize flexible avoidance behaviors if such kinematic diversity is absent from the training dataset.
Conversely, task failures (e.g., failure to lift, Fig.~\ref{fig:robot}, bottom left; or knocking over the can, Fig.~\ref{fig:robot}, bottom middle) arise when guidance gradients directly conflict with task execution. For instance, in lifting tasks, strong repulsive forces effectively cancel out the necessary vertical actuation when the planner fails to identify a feasible lateral detour.

\rev{\subsection{Real Robot Experiment}

We further evaluated the methods on a real Franka Panda robot across two task types: moving-obstacle avoidance and position-preference grasping, where the robot is guided to grasp different parts of a mug, such as the handle or the body. For moving-obstacle avoidance, the three settings vary in the speed and diversity of the obstacle trajectories, and each setting is tested over 20 trials. As shown in Table~\ref{tbl:results}, guided DP often struggles with fast-moving obstacles: it either responds only when the obstacle is already close or produces abrupt evasive motions that interfere with task execution. These results also support our claim that our guidance formulation improves over naive guidance, which uses only repulsion/attraction heuristics and does not explicitly account for the execution of the underlying task. The position-preference experiment further demonstrates that our framework can support inference-time objectives beyond obstacle avoidance, enabling the robot to grasp different parts of the mug according to the specified preference. In addition, we observe the same trend as reported in SFP~\cite{jiang2025streaming}: streaming policies tend to produce smoother trajectories than diffusion-policy-style generation.

\begin{table}[t]
\centering
\footnotesize
\caption{\textbf{Real Robot Results (Success Rate).}}
\setlength{\tabcolsep}{4pt}
\begin{tabular}{l|ccc|c}
\toprule
& & Moving Obstacles & & Position \\
Method & Slow-Steady & Slow-Diverse & Fast & Preference \\
\midrule
DP    & 65 & 25 &  5 & 85 \\
\midrule
SSIP-Na\"ive & 40 & 65 & 55 & 90 \\
SSIP-STEG  & \textbf{90} & \textbf{75} & 55 & \textbf{95} \\
SSIP-CCG   & \textbf{90} & \textbf{75} & \textbf{65} & 90 \\
\bottomrule
\end{tabular}
\label{tbl:results}
\end{table}

\begin{figure}
    \centering
    \includegraphics[width=1\linewidth]{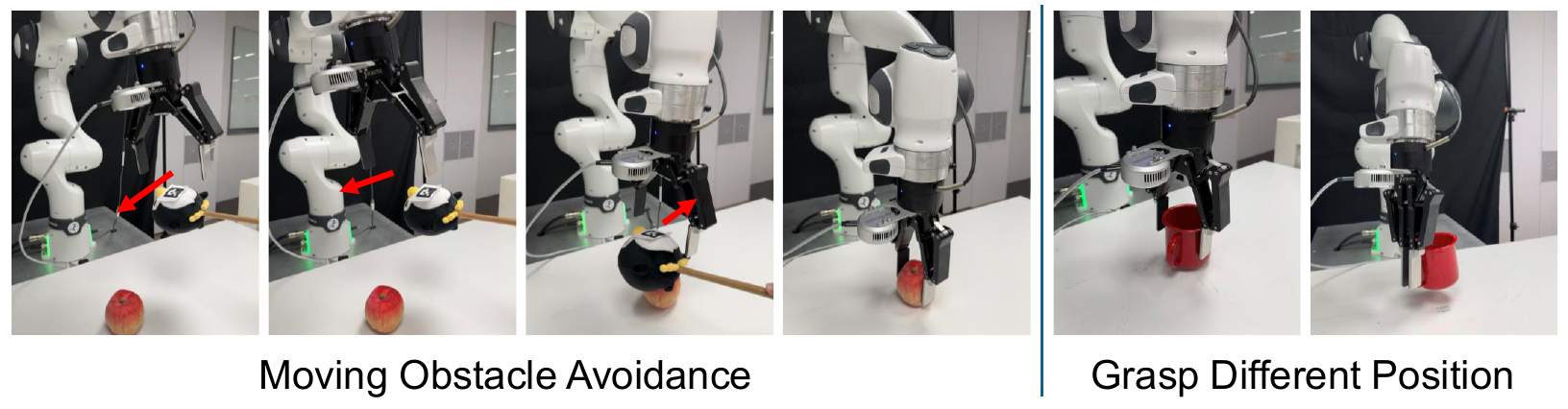}
    \caption{Real Robot Experiment on moving-obstacle avoidance and position-preference grasping.}
    \label{fig:real}
\end{figure}

}

\section{Conclusion, Limitations, and Future Work}
\label{sec:conclusion}
This paper presented a principled framework for inference-time steering of generative robot policies together with a streaming execution substrate. We derived the optimal guidance law for \emph{Stochastic Interpolants} and instantiated it in a \emph{Streaming Stochastic Interpolant Policy (SSIP)} that generates actions incrementally, avoiding the ``wait-then-act'' bottleneck inherent to chunk-based architectures. Empirically, SSIP improves reactivity relative to chunk-based diffusion and flow-policy baselines across both static and dynamic obstacle settings. The proposed guidance instantiations exhibit complementary strengths: CCG provides low-latency, amortized steering when task and constraint structure is available at training time, whereas STEG enables training-free adaptation via on-the-fly gradient estimation under distribution shift.

Across benchmarks, we find that effective inference-time steering is regime-dependent and shaped jointly by the execution architecture and the objective surrogate. On Robomimic tasks with unmodeled obstacles intersecting nominal demonstrations, simple local repulsion performs poorly, indicating that successful avoidance often requires \emph{task-aware} deviations rather than purely reactive distance penalties. In contrast, SSIP with amortized guidance (CCG-P) substantially improves success, suggesting that history- and cost-conditioned critics can learn structured, goal-preserving avoidance behaviors when representative deployment conditions are observed during training. Training-free STEG also improves over repulsion, but underperforms CCG-P in settings requiring longer-horizon detours, consistent with limitations from finite-horizon rollouts and coarse discretization. Complementing these static-obstacle results, Push-T with dynamic obstacles highlights the brittleness of chunk-based policies under within-horizon constraint changes, while streaming SSIP maintains higher success; notably, STEG provides the strongest robustness in the most adversarial \textit{Chase} regime, consistent with the benefits of online adaptation.

\mypara{Limitations and Future Work.}
We identify two primary limitations. First, the policy is constrained by the support of the training trajectory distribution, which can restrict kinematic flexibility during close-range avoidance. Second, STEG operates over a finite prediction horizon, limiting its ability to anticipate obstacles that require longer-term planning. These effects compound: limited horizon discourages early detours, while distributional constraints hinder late-stage maneuvering.
Future work will focus on extending the effective look-ahead horizon of streaming guidance, for example through longer-horizon rollouts or hierarchical planning interfaces. \del{We also plan to validate the proposed framework on real robotic platforms to assess robustness under physical uncertainty, sensor noise, and perception latency.}

\section*{Acknowledgements}

This research / project is supported by the National Research Foundation, Singapore, under its Thematic Competitive Research Programme 2025 (NRF-T-CRP-2025-0003)


\balance
\bibliographystyle{plainnat}
\bibliography{references}

\clearpage
\nobalance
\appendices
\section{Extended Formulation of SSIP}
\label{app:ssip}

In this section, we provide the rigorous mathematical derivation of the Streaming Stochastic Interpolant Policy (SSIP), extending the formulation presented in Sec.~\ref{sec:method}. We detail how SSIP generalizes the deterministic Streaming Flow Policy (SFP)~\cite{jiang2025streaming} into the Stochastic Interpolant framework~\cite{albergo2023stochastic}, enabling the derivation of the guidance law used in our main results.

\subsection{Deterministic Backbone: Streaming Flow Policy (SFP)}
We begin with the underlying deterministic structure defined by SFP. Unlike standard diffusion policies that generate strictly coupled action chunks, SFP aligns the flow time $t$ with the robot's execution time.
Let $\boldsymbol{\xi} \in \mathbb{R}^{T \times D}$ denote the ground-truth demonstration trajectory (the target data). SFP constructs a probability path concentrated around $\boldsymbol{\xi}_t$, conceptually forming a ``Gaussian Tube'':
\begin{equation}
    p_{SFP}(\mathbf{a}_t | \boldsymbol{\xi}) = \mathcal{N}(\boldsymbol{\xi}_t, \sigma^2(t) \mathbf{I})
\end{equation}
where $\sigma(t) = \sigma_0 e^{-kt}$ describes a contracting variance schedule with stabilizing gain $k > 0$ and initial dispersion $\sigma_0$.
The vector field governing this probability path follows the linear Ordinary Differential Equation (ODE):
\begin{equation}
    \frac{d \mathbf{a}_t}{dt} = \mathbf{v}_{SFP}(\mathbf{a}_t | \boldsymbol{\xi}) = \dot{\boldsymbol{\xi}}_t - k(\mathbf{a}_t - \boldsymbol{\xi}_t)
\end{equation}
In SFP, a velocity network $v_\theta(\mathbf{a}_t, t)$ is trained to regress this target field. However, the resulting dynamics are fundamentally deterministic, precluding the direct application of score-based guidance which requires a diffusive term.

\subsection{Stochastic Interpolation Mechanism}
SSIP upgrades this deterministic backbone into a Stochastic Interpolant. Consistent with the definition in Eq.~(1) of the main text, we define the stochastic process $\mathbf{a}_t$ as a superposition of the deterministic interpolant $I(t, \cdot)$ and an independent noise process:
\begin{equation}
    \mathbf{a}_t = \underbrace{\boldsymbol{\xi}_t + \sigma(t) \boldsymbol{\epsilon}}_{I(t, \dots) \text{ (SFP Backbone)}} + \underbrace{\gamma(t) \mathbf{z}}_{\text{SI Noise}}
    \label{eq:ssip_interpolant_detailed}
\end{equation}
where:
\begin{itemize}
    \item $\boldsymbol{\epsilon} \sim \mathcal{N}(\mathbf{0}, \mathbf{I})$ represents the variance of the underlying valid behavior (inherited from SFP).
    \item $\mathbf{z} \sim \mathcal{N}(\mathbf{0}, \mathbf{I})$ is the auxiliary noise introduced by SSIP for diffusive sampling.
    \item $\gamma(t)$ is the interpolant noise schedule (e.g., $\gamma(t) \propto \sqrt{t(1-t)}$), distinct from the flow variance $\sigma(t)$.
\end{itemize}
This formulation effectively ``inflates'' the deterministic Gaussian tube into a stochastic probability cloud, admitting a valid score function $\nabla \log \rho_t(\mathbf{a}_t)$ required for our guidance derivation.

\subsection{Training Objectives}
To realize the SDE-based inference, we estimate two fields: the transport velocity $v_\theta$ and the score function $\mathbf{s}_\theta$.

\subsubsection{Velocity Matching}
The velocity network $v_\theta(\mathbf{a}_t, t)$ learns the stabilizing vector field of the underlying flow. It is trained to regress the SFP drift even when the input state is perturbed by the additional SI noise:
\begin{equation}
    \mathcal{L}_{vel}(\theta) = \mathbb{E}_{t, \boldsymbol{\xi}, \boldsymbol{\epsilon}, \mathbf{z}} \left[ \| v_\theta(\mathbf{a}_t, t) - \mathbf{v}_{SFP}(\mathbf{a}_t^{SFP} | \boldsymbol{\xi}) \|^2 \right]
\end{equation}
where $\mathbf{a}_t^{SFP} = \boldsymbol{\xi}_t + \sigma(t)\boldsymbol{\epsilon}$ is the unperturbed SFP state.

\subsubsection{Score Matching}
The denoiser network $\eta_\theta(\mathbf{a}_t, t)$ estimates the added SI noise $\mathbf{z}$. This is trained via standard denoising score matching:
\begin{equation}
    \mathcal{L}_{score}(\theta) = \mathbb{E}_{t, \mathbf{a}_t, \mathbf{z}} \left[ \| \eta_\theta(\mathbf{a}_t, t) - \mathbf{z} \|^2 \right]
\end{equation}
The analytical score of the conditional distribution is then given by $\mathbf{s}_\theta(\mathbf{a}_t, t) \approx -\frac{\eta_\theta(\mathbf{a}_t, t)}{\gamma(t)}$. 

We avoid directly regressing the score function because it scales with $1/\gamma(t)$, which leads to numerical instability as $t \to 0$ or $t \to 1$. Instead, we adopt the noise-prediction parameterization proposed in the original SI framework~\cite{albergo2023stochastic} and utilized in~\cite{chen24bridger}, which ensures stable training objectives.

\subsection{Inference via Stochastic Differential Equation}
As introduced in Eq.~(2) of the main text, the marginal density of the SSIP process satisfies the Fokker-Planck equation associated with a generic SDE. By combining the learned velocity and score, we derive the specific form of the drift $\mathbf{b}(\mathbf{a}_t, t)$ for SSIP:
\begin{equation}
    d\mathbf{a}_t = \mathbf{b}(\mathbf{a}_t, t) dt + \sqrt{2\epsilon(t)} d\mathbf{W}_t
\end{equation}
The drift $\mathbf{b}$ that preserves the marginals of Eq.~\eqref{eq:ssip_interpolant_detailed} is~\cite{albergo2023stochastic}:
\begin{equation}
    \mathbf{b}(\mathbf{a}_t, t) = v_\theta(\mathbf{a}_t, t) + \underbrace{\left( \epsilon(t) - \gamma(t)\dot{\gamma}(t) \right)}_{\text{Diffusivity Correction}} \mathbf{s}_\theta(\mathbf{a}_t, t)
\end{equation}
Here, $\epsilon(t)$ is the user-defined diffusivity schedule. This formulation is the theoretical foundation for our guidance methods (STEG and CCG), as the presence of the diffusion term $\sqrt{2\epsilon(t)} d\mathbf{W}_t$ and the score term $\mathbf{s}_\theta$ allows for the injection of the guidance gradient $\nabla J$ as a bias to the score.

\section{Theoretical Analysis of Test-Time Estimation}
\label{app:estimator_theory}

In this section, we provide a rigorous error decomposition of the guidance gradient estimated by STEG. We quantify the gap between our estimator and the true physical oracle, justifying our design choices of using a differentiable ensemble (STEG) and the ``Large-$N$, Small-$K$" strategy.

\subsection{Unified Formulation: The Posterior Force}
We adopt a \textbf{pathwise perspective} based on the \textbf{reparameterization trick} to derive the guidance law. Let the true physical dynamics be represented by the deterministic mapping $\tau = \Phi_{phy}(\mathbf{a}, \boldsymbol{\epsilon})$, where $\mathbf{a}$ is the control action and $\boldsymbol{\epsilon} \sim p(\boldsymbol{\epsilon})$ represents the inherent physical stochasticity.

\subsubsection{The Physical Oracle ($S^*$)}
We define the optimal guidance score $S^*(\mathbf{a})$ as the gradient of the log-expected utility under the true physics. This formulation aligns with the derivation in Section~\ref{guide_streaming}, omitting the scaling factor for simplicity:
\begin{equation}
    S^*(\mathbf{a}) = \nabla_{\mathbf{a}} \log \mathbb{E}_{\boldsymbol{\epsilon}} \left[ e^{-J(\Phi_{phy}(\mathbf{a}, \boldsymbol{\epsilon}))} \right]
\end{equation}

By applying the \textbf{reparameterization trick}, the expectation is taken over the fixed distribution of $\boldsymbol{\epsilon}$, while the gradient is computed with respect to $\mathbf{a}$. This allows us to move the gradient operator inside the expectation. We first apply the log-derivative identity $\nabla \log Z = \frac{\nabla Z}{Z}$:
\begin{equation}
    S^*(\mathbf{a}) = \frac{\nabla_{\mathbf{a}} \mathbb{E}_{\boldsymbol{\epsilon}} \left[ e^{-J(\Phi_{phy}(\mathbf{a}, \boldsymbol{\epsilon}))} \right]}{\mathbb{E}_{\boldsymbol{\epsilon}} \left[ e^{-J(\Phi_{phy}(\mathbf{a}, \boldsymbol{\epsilon}))} \right]}
    \label{eq:oracle_fraction}
\end{equation}

Next, we apply the chain rule to the term $e^{-J(\tau)}$:
\begin{equation}
    \begin{aligned}
    \nabla_{\mathbf{a}} e^{-J(\tau)} &= e^{-J(\tau)} \cdot \left( - \nabla_{\mathbf{a}} J(\tau) \right) \\
    &= e^{-J(\tau)} \underbrace{\left( - \nabla_\tau J(\tau) \cdot \frac{\partial \Phi_{phy}}{\partial \mathbf{a}} \right)}_{\text{Backpropagated Force}}
    \end{aligned}
\end{equation}
Substituting this back into the numerator of Eq.~\eqref{eq:oracle_fraction}:
\begin{equation}
    S^*(\mathbf{a}) = \frac{\mathbb{E}_{\boldsymbol{\epsilon}} \left[ e^{-J(\tau)} \left( -\nabla_\tau J(\tau) \cdot \mathbf{J}_{phy}(\tau) \right) \right]}{\mathbb{E}_{\boldsymbol{\epsilon}} \left[ e^{-J(\tau)} \right]}
\end{equation}
where $\mathbf{J}_{phy}(\tau) = \frac{\partial \Phi_{phy}}{\partial \mathbf{a}}$ is the Jacobian of the physical dynamics.

To interpret this geometrically, we define the \textbf{Physical Safe Posterior} distribution $p_{safe}(\tau|\mathbf{a})$. This corresponds to the physical distribution $p(\tau|\mathbf{a})$ re-weighted by the safety utility:
\begin{equation}
    p_{safe}(\tau|\mathbf{a}) \coloneqq \frac{p(\tau|\mathbf{a}) e^{-J(\tau)}}{\underbrace{\mathbb{E}_{\tau' \sim p} [e^{-J(\tau')}]}_{\text{Normalization Constant } Z}}
\end{equation}
Observing that the denominator in our score equation is exactly the normalization constant $Z$, we can rewrite the oracle gradient as an expectation under this safe posterior:
\begin{equation}
    S^*(\mathbf{a}) = \mathbb{E}_{\tau \sim p_{safe}(\cdot|\mathbf{a})} \left[ \mathcal{F}_{phy}(\tau, \mathbf{a}) \right]
\end{equation}
Here, $\mathcal{F}_{phy}(\tau, \mathbf{a}) \triangleq -\nabla_\tau J(\tau) \cdot \mathbf{J}_{phy}(\tau)$ represents the \textit{Physical Repulsive Force} for a single trajectory—the vector indicating how to adjust $\mathbf{a}$ to reduce the collision cost $J$ given a specific physical realization $\boldsymbol{\epsilon}$.

\subsubsection{The Estimator Gradient ($\hat{S}$)}
Since the ground truth physical oracle is inaccessible, STEG approximates future trajectories. We define an estimator distribution $q(\tau|\mathbf{a})$ via the learned differentiable surrogate $\tau = \Phi_{est}(\mathbf{a}, \mathbf{z})$, where $\mathbf{z}$ is the noise injected during the SDE rollout.
Analogous to the physical derivation, the STEG gradient (computed via LogSumExp in our implementation) mathematically approximates the expectation of the modeled force under the estimator's safe posterior $q_{safe}$:
\begin{equation}
    \hat{S}(\mathbf{a}) = \mathbb{E}_{\tau \sim q_{safe}(\cdot|\mathbf{a})} \left[ \mathcal{F}_{est}(\tau, \mathbf{a}) \right]
\end{equation}
where $\mathcal{F}_{est} \triangleq -\nabla_\tau J(\tau) \cdot \mathbf{J}_{est}(\tau)$ uses the Jacobian of the learned velocity network $\mathbf{J}_{est} = \frac{\partial \Phi_{est}}{\partial \mathbf{a}}$.

\subsection{Rigorous Error Decomposition}
To quantify the estimation error $\Delta S(\mathbf{a}) = \hat{S}(\mathbf{a}) - S^*(\mathbf{a})$, we add and subtract the intermediate term $\mathbb{E}_{p_{safe}}[\mathcal{F}_{est}]$ (the estimated force evaluated on true physical trajectories). This decomposes the error into two structurally distinct sources:

\begin{equation}
    \label{eq:error_decomp_full}
    \begin{split}
    \Delta S(\mathbf{a}) &= \mathbb{E}_{q_{safe}}[\mathcal{F}_{est}] - \mathbb{E}_{p_{safe}}[\mathcal{F}_{phy}] \\
    &= \underbrace{\left( \mathbb{E}_{q_{safe}}[\mathcal{F}_{est}] - \mathbb{E}_{p_{safe}}[\mathcal{F}_{est}] \right)}_{\text{Term I: Distributional Shift}} \\
    &\quad + \underbrace{\mathbb{E}_{p_{safe}} \left[ \mathcal{F}_{est} - \mathcal{F}_{phy} \right]}_{\text{Term II: Dynamics Mismatch}}
    \end{split}
\end{equation}

\subsubsection{Term I: Distributional Shift (Coverage Error)}
Term I measures the support mismatch between the estimator $q$ and the physics $p$.
\begin{itemize}
    \item \textbf{Problem:} If the physical safety landscape is multi-modal (e.g., passing an obstacle on the left or right) but the estimator $q$ is too narrow or uses a single deterministic rollout, $q_{safe}$ may collapse to a single mode or miss the safe region entirely.
    \item \textbf{STEG Solution (Large-$N$):} STEG minimizes this by generating a large ensemble ($N \ge 30$) with stochastic noise injection. This ensures that the support of $q$ asymptotically covers the high-probability regions of $p_{safe}$, driving Term I to zero.
\end{itemize}

\subsubsection{Term II: Dynamics Mismatch (Jacobian Error)}
Term II measures the alignment error between the modeled and physical dynamics, weighted by the safety of the trajectory.
\begin{itemize}
    \item \textbf{Problem:} Even if the trajectories are kinematically similar, if the Jacobians diverge ($\mathbf{J}_{est} \neq \mathbf{J}_{phy}$), the guidance force will point in a physically invalid direction (e.g., commanding instantaneous lateral motion for a non-holonomic robot).
    \item \textbf{STEG Solution (Differentiable Physics):} By unrolling the Euler integration of the learned velocity field $v_\theta$, STEG enforces structural similarity between $\Phi_{est}$ and $\Phi_{phy}$. This ensures that $\mathbf{J}_{est} \approx \mathbf{J}_{phy}$, minimizing Term II.
\end{itemize}

\subsection{Practical Strategy: Large-$N$, Small-$K$}
Based on this decomposition, we justify the specific hyperparameters used in STEG. The estimation error is practically dominated by Selection Error (finite samples) and Jacobian Drift (long-horizon divergence).

\paragraph{1. Large-$N$ for Selection Error ($\epsilon_{sel}$)}
The safety posterior $p_{safe}$ effectively acts as a Softmax gating function. To accurately estimate the expectation $\mathbb{E}_{q_{safe}}$, the ensemble must contain at least a few "survivor" particles that have low cost. 
For a cluttered environment with success probability $\rho \ll 1$, the probability of ensemble failure is $(1-\rho)^N$. By utilizing SIMD parallelism to scale $N$ (Large-$N$), we exponentially decay this selection error without increasing wall-clock latency.

\paragraph{2. Small-$K$ for Low Latency (Efficiency vs. Accuracy)}
Since the rollout steps must be executed sequentially, the horizon length $K$ constitutes the primary bottleneck for wall-clock inference time. Increasing $K$ linearly increases latency, potentially compromising the robot's ability to react to sudden changes.

To ensure real-time performance (e.g., control frequencies $>20$ Hz), STEG adopts a \textbf{``Small-$K$, Large-$\Delta t$"} strategy (e.g., $K=3/4$).
While a larger integration step $\Delta t$ inevitably introduces some discretization error (a form of dynamics mismatch), we observe that the guidance gradient for obstacle avoidance is robust to this approximation. The repulsive force is dominated by the relative geometry of impending collisions; a coarse trajectory that correctly captures the \textit{existence} of a collision is sufficient to yield a valid avoidance direction. Thus, we prioritize minimizing latency over minimizing integration error, as delayed reactions are far more detrimental to safety than slightly approximate gradients.

\section{Guidance Implementation Details}

We provide implementation details for our two proposed guidance mechanisms: the training-free Stochastic Trajectory Ensemble Guidance (STEG) and the amortized Conditional Critic Guidance (CCG).

\subsection{Stochastic Trajectory Ensemble Guidance (STEG)}
STEG computes guidance gradients on-the-fly via parallel simulations. The hyperparameters are selected to balance the trade-off between gradient variance and inference latency (as analyzed in Appendix B):
\begin{itemize}
    \item \textbf{Ensemble Size ($N$):} We employ an ensemble of $N$ parallel trajectories. This size strikes a balance between covering diverse modes in the action distribution and maintaining computational efficiency.
    \begin{itemize}
        \item Push-T: $N=64$
        \item Robomimic: $N=32$
    \end{itemize}
    
    \item \textbf{Rollout Horizon ($K$):} We adopt a short-horizon strategy (denoted as $K$) to minimize inference latency.
    \begin{itemize}
        \item Push-T: $K=3$ steps 
        \item Robomimic: $K=4$ steps 
    \end{itemize}
    
    \item \textbf{Integration Step ($\Delta t$):} We utilize a coarse discretization step size of $\Delta t$  to facilitate efficient gradient estimation during the rollout phase.
    \begin{itemize}
        \item Push-T: $\Delta t=0.15$
        \item Robomimic: $\Delta t=0.1$
    \end{itemize}
\end{itemize}

\subsection{Conditional Critic Guidance (CCG)}
To reduce inference latency, CCG amortizes the rollout computation into a learned value function $V_\psi$.

\subsubsection{Network Architecture}
We employ a ResNet-based critic with 6 hidden layers of size 1024. The network $V_\psi(\mathbf{a}_t, t, \mathbf{h}, \phi)$ conditions on the current action $\mathbf{a}_t$, diffusion time $t$, the history embedding $\mathbf{h}$ (extracted from the frozen policy encoder), and obstacle parameters $\phi$.

\subsubsection{Training Objective}
We train the critic using a regression approach anchored by Monte Carlo rollouts collected from the frozen base policy. For a sampled tuple $(\mathbf{a}_t, \mathbf{h}, \phi)$, we simulate $K$ parallel future trajectories $\{\tau^{(k)}\}_{k=1}^K$ using the base SSIP dynamics and compute an empirical target value $y_{target}$. The network minimizes the $L_2$ regression loss:
\begin{equation}
    \mathcal{L} = \mathbb{E} \left[ \| V_\psi(\mathbf{a}_t, t, \mathbf{h}, \phi) - y_{target} \|^2 \right]
\end{equation}

\subsubsection{Cost Variants}
We instantiate two variants of CCG by defining different regression targets $y_{target}$ based on the rollout outcomes:
\begin{itemize}
    \item \textbf{CCG-D (Distance Potential):} The target approximates the log-expected future utility, $y_{target} \approx \log \mathbb{E}[e^{-J(\tau)}]$. The trajectory cost is defined as the cumulative distance potential: $J(\tau) = \sum \exp\left(-\frac{\|x - x_{obs}\|^2}{2\sigma^2}\right)$.
    \item \textbf{CCG-P (Collision Proxy):} The target $y_{target}$ is a probabilistic collision risk computed from the ensemble's statistics. 
    \begin{itemize}
        \item For \textbf{3D manipulation tasks} (Robomimic), we employ a hybrid risk metric combining minimum distance and approach alignment (aiming probability) to encourage early avoidance: $y_{target} = \max(P_{collision}, \lambda \cdot P_{aiming})$.
        \item For the \textbf{2D Push-T task}, we utilize purely the collision probability ($y_{target} = P_{collision}$) derived from the minimum distance. Empirical analysis revealed that the aiming term in this highly dynamic task introduced gradient instability without improving avoidance success.
    \end{itemize}
    This variant relaxes the strict additive-cost assumption of the Doob-h transform but provides a practical, highly reactive guidance field for dynamic obstacles.
\end{itemize}

\clearpage
\begin{algorithm*}
    \caption{SSIP Inference with Stochastic Trajectory Ensemble Guidance (STEG)}
    \label{alg:steg_inference}
    \begin{algorithmic}[1]
        \REQUIRE SSIP Policy $\pi_\theta$, Ensemble Size $N$, Horizon $K$, Obstacle $\phi$, Scale $\lambda$, Activation Dist $d_{act}$, Policy Horizon $H$, Execution Horizon $H_{exec}$
        
        \STATE \COMMENT{--- Initialization ---}
        \STATE Get initial observation $\mathbf{o}_{init}$ from environment
        \STATE Initialize latent action trajectory $\mathbf{a}_0$ from $\mathbf{o}_{init}$
        \STATE $\Delta t \leftarrow 1/H$ \COMMENT{Step size for flow integration}
        
        \FOR{env step $i = 0$ to $N_{max}$}
            \STATE Get observation $\mathbf{o}$ from environment
            \STATE $t \leftarrow \text{Clip}((i \pmod {H_{exec}}) \cdot \Delta t, 0, 1)$ \COMMENT{Flow time aligned with execution}
            
            \STATE \COMMENT{--- 1. Base Flow Drift ---}
            \STATE Predict velocity: $\mathbf{v}_\theta \leftarrow \pi_\theta.\text{velocity}(\mathbf{a}_t, t, \mathbf{o})$
            \STATE Predict score: $\mathbf{s}_\theta \leftarrow \pi_\theta.\text{score}(\mathbf{a}_t, t, \mathbf{o})$
            \STATE Compute base drift: $\mathbf{b}_{base} \leftarrow \mathbf{v}_\theta - (\gamma(t) \dot{\gamma}(t)) \mathbf{s}_\theta$ \COMMENT{Probability Flow ODE}
            
            \STATE \COMMENT{--- 2. Stochastic Ensemble Guidance ---}
            \STATE Calculate distance: $d_{obs} \leftarrow \|\text{Denormalize}(\mathbf{a}_t) - \phi\|$
            \IF{$d_{obs} < d_{act}$}
                \STATE Enable gradient: $\mathbf{a}_{in} \leftarrow \mathbf{a}_t.\text{requires\_grad}()$
                \STATE \textbf{Ensemble Rollout:} $\mathbf{A}^{(0)} \leftarrow \text{Repeat}(\mathbf{a}_{in}, N)$
                \STATE Initialize cost: $\mathbf{C} \leftarrow \mathbf{0} \in \mathbb{R}^N$
                
                \FOR{$k = 0$ to $K-1$}
                    \STATE $t' \leftarrow t + k \Delta t_{sim}$
                    \STATE Predict drift: $\mathbf{B}_{drift} \leftarrow \pi_\theta(\mathbf{A}^{(k)}, t', \mathbf{o})$
                    \STATE SDE Step: $\mathbf{A}^{(k+1)} \leftarrow \mathbf{A}^{(k)} + \mathbf{B}_{drift} \Delta t_{sim} + \sigma \sqrt{\Delta t_{sim}} \mathbf{Z}$
                    \STATE Accumulate Cost: $\mathbf{C} \leftarrow \mathbf{C} + J(\mathbf{A}^{(k+1)}, \phi) \cdot \Delta t_{sim}$
                \ENDFOR
                
                \STATE Utility: $U \leftarrow -\mathbf{C}$
                \STATE Log-Sum-Exp: $L \leftarrow \log \sum_{n=1}^N \exp(U_n)$ \COMMENT{Soft-max over ensemble trajectories}
                \STATE Compute Gradient: $\mathbf{g}_{teg} \leftarrow \nabla_{\mathbf{a}} L$
                \STATE \textbf{Dynamic Weighting:} $w_{dyn} \leftarrow \lambda \cdot (1 - d_{obs}/d_{act})$ \COMMENT{Linear ramp-up based on proximity}
                \STATE Guided Drift: $\mathbf{b}_{total} \leftarrow \mathbf{b}_{base} + w_{dyn} \cdot \mathbf{g}_{teg}$
            \ELSE
                \STATE $\mathbf{b}_{total} \leftarrow \mathbf{b}_{base}$
            \ENDIF
            
            \STATE \COMMENT{--- 3. Streaming Integration ---}
            \STATE Update trajectory: $\mathbf{a}_{next} \leftarrow \mathbf{a}_t + \mathbf{b}_{total} \cdot \Delta t$ 
            \STATE $\mathbf{a}_{phys} \leftarrow \text{Denormalize}(\mathbf{a}_{next})$
            \STATE Execute $\mathbf{a}_{phys}$ in environment
            
            \STATE \COMMENT{--- 4. Streaming Handoff ---}
            \IF{$(i+1) \pmod {H_{exec}} == 0$}
                \STATE $\mathbf{a}_{t=0} \leftarrow \text{Re-Initialize}(\mathbf{o})$ \COMMENT{Shift to next chunk}
            \ELSE
                \STATE $\mathbf{a}_t \leftarrow \mathbf{a}_{next}$ \COMMENT{Pass updated trajectory to next step}
            \ENDIF
        \ENDFOR
    \end{algorithmic}
\end{algorithm*}
\clearpage

\clearpage
\begin{algorithm*}
    \caption{SSIP Inference with Conditional Critic Guidance (CCG)}
    \label{alg:ccg_inference}
    \begin{algorithmic}[1]
        \REQUIRE SSIP Policy $\pi_\theta$, Critic $V_\psi$, Obstacle $\phi$, Scale $\lambda$, Power $k$, Policy Horizon $H$, Execution Horizon $H_{exec}$
        
        \STATE \COMMENT{--- Initialization ---}
        \STATE Get initial observation $\mathbf{o}_{init}$ from environment
        \STATE Initialize latent action trajectory $\mathbf{a}_0$ from $\mathbf{o}_{init}$
        \STATE $\Delta t \leftarrow 1/H$ \COMMENT{Step size for flow integration}
        
        \FOR{env step $i = 0$ to $N_{max}$}
            \STATE Get observation $\mathbf{o}$ from environment
            \STATE $t \leftarrow \text{Clip}((i \pmod {H_{exec}}) \cdot \Delta t, 0, 1)$ \COMMENT{Flow time aligned with execution}
            
            \STATE \COMMENT{--- 1. Base Flow Drift ---}
            \STATE Predict velocity: $\mathbf{v}_\theta \leftarrow \pi_\theta.\text{velocity}(\mathbf{a}_t, t, \mathbf{o})$
            \STATE Predict score: $\mathbf{s}_\theta \leftarrow \pi_\theta.\text{score}(\mathbf{a}_t, t, \mathbf{o})$
            \STATE Compute base drift: $\mathbf{b}_{base} \leftarrow \mathbf{v}_\theta - (\gamma(t) \dot{\gamma}(t)) \mathbf{s}_\theta$ \COMMENT{Probability Flow ODE}
            
            \STATE \COMMENT{--- 2. Adaptive Critic Guidance ---}
            \IF{$\lambda > 0$}
                \STATE Enable gradient: $\mathbf{a}_{in} \leftarrow \mathbf{a}_t.\text{requires\_grad}()$
                \STATE Predict Risk Logit: $v_{risk} \leftarrow V_\psi(\mathbf{a}_{in}, t, \mathbf{o}, \phi)$
                \STATE Collision Probability: $p_{risk} \leftarrow \sigma(v_{risk})$
                \STATE \textbf{Soft Gating:} $w_{dyn} \leftarrow \lambda \cdot (p_{risk})^k$ \COMMENT{Power scaling suppresses guidance in safe states}
                \STATE Compute Gradient: $\mathbf{g}_{v} \leftarrow \nabla_{\mathbf{a}} v_{risk}$ \COMMENT{Gradient w.r.t Logit value}
                \STATE Guided Drift: $\mathbf{b}_{total} \leftarrow \mathbf{b}_{base} - w_{dyn} \cdot \mathbf{g}_{v}$
            \ELSE
                \STATE $\mathbf{b}_{total} \leftarrow \mathbf{b}_{base}$
            \ENDIF
            
            \STATE \COMMENT{--- 3. Streaming Integration ---}
            \STATE Update trajectory: $\mathbf{a}_{next} \leftarrow \mathbf{a}_t + \mathbf{b}_{total} \cdot \Delta t$ 
            \STATE $\mathbf{a}_{phys} \leftarrow \text{Denormalize}(\mathbf{a}_{next})$
            \STATE Execute $\mathbf{a}_{phys}$ in environment
            
            \STATE \COMMENT{--- 4. Streaming Handoff ---}
            \IF{$(i+1) \pmod H_{exec} == 0$}
                \STATE $\mathbf{a}_{t=0} \leftarrow \text{Re-Initialize}(\mathbf{o})$ \COMMENT{Shift to next chunk}
            \ELSE
                \STATE $\mathbf{a}_t \leftarrow \mathbf{a}_{next}$ \COMMENT{Pass updated trajectory to next step}
            \ENDIF
        \ENDFOR
    \end{algorithmic}
\end{algorithm*}
\clearpage

\begin{table*}[t!] 
    \centering
    \caption{Hyperparameters for Policy Training}
    \label{tab:hyperparams_policy}
    \setlength{\tabcolsep}{12pt} 
    \begin{tabular}{l|ccc}
        \toprule
        \textbf{Parameter} & \textbf{SSIP (Ours)} & \textbf{Diffusion Policy (DP)} & \textbf{Flow Policy (FP)} \\
        \midrule
        Backbone Architecture & ConditionalUnet1D & ConditionalUnet1D & ConditionalUnet1D \\
        Hidden Dimensions & [256, 512, 1024] & [256, 512, 1024] & [256, 512, 1024] \\
        Optimizer & AdamW & AdamW & AdamW \\
        Learning Rate & 1e-4 & 1e-4 & 1e-4 \\
        Batch Size & 64 & 64 & 64 \\
        Action Horizon ($H$) & 16 & 16 & 16 \\
        Execution Horizon ($H_{exec}$) & 8 (Streaming) & 8 (Chunking) & 8 (Chunking) \\
        Noise Schedule $\gamma(t)$ & $0.1\sqrt{t(1-t)}$ & Squared Cosine & $\sigma_0 \exp(-kt)$ \\
        Training Epochs & 1200 & 1200 & 1200 \\
        \bottomrule
    \end{tabular}
\end{table*}

\section{Experimental Details}
\label{app:experimental_details}

In this section, we provide detailed hyperparameters, environment configurations, and training procedures to ensure reproducibility.

\subsection{Environment and Task Setup}
We evaluate our method on two domains: the 2D planar Push-T task and the 3D Robomimic manipulation suite.

\subsubsection{Push-T Task}
The goal is to push a T-shaped block to a target pose. The state space consists of the robot end-effector position and the block pose (position and angle).
\begin{itemize}
    \item \textbf{Observation Space:} Low-dimensional state vector, including agent position $(x, y)$ and block pose $(x, y, \theta)$.
    \item \textbf{Action Space:} Continuous control space (2D position for Push-T; End-effector pose for Robomimic). The model outputs the drift term used to evolve the action state $a_t$.
    \item \textbf{Simulation Horizon:} $T = 250$ steps.
    \item \textbf{Control Frequency:} 10 Hz.
    \item \textbf{Obstacle Configuration:} We introduce both static and dynamic obstacles to rigorously test reactivity.
    \begin{itemize}
        \item \textbf{Static Obstacles:} 8 circular obstacles are placed at fixed locations to form a cluttered environment.
        \item \textbf{Dynamic Obstacles:} We implement three distinct dynamic behaviors to evaluate different aspects of avoidance:
        \begin{enumerate}
            \item \textbf{Active Interception:} The obstacle is initialized at an offset from a future waypoint on the reference trajectory. It moves linearly to intercept the target point with a fixed velocity profile designed to reach the interception point within $T=50$ steps, effectively creating a ``moving roadblock" that forces the agent to brake or detour. Once the target is reached (or time limit exceeded), the obstacle halts to block the path.
            
            \item \textbf{Sinusoidal Oscillation:} The obstacle oscillates perpendicular to the nominal trajectory anchor point. Its position $\mathbf{p}(t)$ evolves as:
            \begin{equation}
                \mathbf{p}(t) = \mathbf{p}_{anchor} + A \cdot \sin(2\pi f t) \cdot \mathbf{v}_{\perp}
            \end{equation}
            where amplitude $A=40$ pixels, frequency $f \approx 0.03$ (normalized step frequency), and $\mathbf{v}_{\perp}$ is the unit vector perpendicular to the trajectory tangent.
            
            \item \textbf{Chasing:} The obstacle actively pursues the agent's current position $\mathbf{p}_{agent}$. The velocity is updated using a proportional controller with inertia:
            \begin{equation}
                \mathbf{v}_{t+1} = (1 - \alpha)\mathbf{v}_t + \alpha \cdot v_{max} \frac{\mathbf{p}_{agent} - \mathbf{p}_t}{\|\mathbf{p}_{agent} - \mathbf{p}_t\|}
            \end{equation}
            where $\alpha=0.4$ is the turn rate (simulating inertia) and $v_{max}=3.0$ is the chase speed.
        \end{enumerate}
    \end{itemize}
\end{itemize}

\subsubsection{Robomimic Tasks}
We use the standard Lift, Can, and Square tasks from the Robomimic suite.
\begin{itemize}
    \item \textbf{Observation Space:} Low-dimensional state vector consisting of the object state and proprioceptive states (e.g., end-effector pose and gripper joint positions).
    \item \textbf{Action Space:} Continuous control space (End-effector pose for Robomimic).
    \item \textbf{Simulation Horizon:} $T = 400$ steps.
    \item \textbf{Control Frequency:} 20 Hz.
    \item \textbf{Obstacle Configuration:} Unmodeled static obstacles are placed to intersect the nominal demonstration trajectories, forcing the agent to deviate.
\end{itemize}

\subsection{Policy Architecture and Training}
We compare our Streaming Stochastic Interpolant Policy (SSIP) against Diffusion Policy (DP) and Flow Policy (FP). Table~\ref{tab:hyperparams_policy} summarizes the hyperparameters for the Robomimic experiments. For the Push-T experiments, we trained SSIP and FP from scratch, while utilizing the fully converged DP checkpoints provided by \cite{jiang2025streaming}. All policies employ the same backbone architecture and parameters as detailed in Table~\ref{tab:hyperparams_policy} and were trained to convergence, achieving high rewards.

\subsection{Baseline Implementation Details}
\label{sec:baseline_details}

To ensure a fair comparison, we implemented two categories of baseline guidance mechanisms: Gradient-based Conventional Guidance (for DP and FP) and Naive Guidance based on analytical potential fields (for our SSIP).

\subsubsection{Diffusion Policy with Reconstruction Guidance \cite{janner2022planning}}
This method follows the classifier-guidance paradigm, utilizing the intermediate diffusion state $x_t$ to estimate the clean trajectory $x_0$ for gradient calculation.

\paragraph{Reconstruction \& Gradient Injection.}
At each reverse diffusion step $t$, we estimate the clean data $\hat{x}_0$ from the current noisy sample $x_t$ and the noise prediction $\epsilon_\theta(x_t)$ using Tweedie's formula:
\begin{equation}
    \hat{x}_0 = \frac{x_t - \sqrt{1 - \bar{\alpha}_t} \epsilon_\theta(x_t)}{\sqrt{\bar{\alpha}_t}}
\end{equation}
We define a distance-based cost function $J(\hat{x}_0)$ on this estimated trajectory. The gradient $\nabla_{x_t} J(\hat{x}_0)$ is computed via backpropagation and injected into the noise prediction:
\begin{equation}
    \hat{\epsilon} = \epsilon_\theta(x_t) + w(t) \sqrt{1 - \bar{\alpha}_t} \nabla_{x_t} J(\hat{x}_0)
\end{equation}
We observed that $\hat{x}_0$ estimates are inaccurate during the early stages of generation. Therefore, we employ a time-dependent weight $w(t)$ scaled by the signal-to-noise ratio to prevent incorrect guidance early in the process.

\subsubsection{Flow Policy  Guidance \cite{feng2025guidance}}
For the flow matching policy, we employ a lookahead mechanism that linearly extrapolates the future state using the current velocity field.

\paragraph{Lookahead Extrapolation.}
During the ODE integration step $t \in [0, 1]$, we predict the terminal state $\hat{x}_1$ using the current velocity $v_\theta(x_t, t)$:
\begin{equation}
    \hat{x}_1 = x_t + v_\theta(x_t, t) \cdot (1 - t)
\end{equation}
The gradient of the cost function $\nabla_{x_t} J(\hat{x}_1)$ is used to modify the velocity field directly:
\begin{equation}
    \mathrm{d}x_t = \left( v_\theta(x_t, t) - \lambda(t) \cdot \nabla_{x_t} J(\hat{x}_1) \right) \mathrm{d}t
\end{equation}
Due to the numerical instability of gradients near obstacles, explicit gradient normalization (clamping) is essential for stability.

\subsubsection{Naive Repulsive Guidance (SSIP)}
As a non-learning baseline, this method applies an analytical repulsive potential field directly to the action space without backpropagation. We apply this to the drift term of the SSIP.

\paragraph{Repulsive Force}
We define a repulsive vector based solely on the geometric distance $d$ between the current trajectory point $x_t$ and the obstacle $x_{obs}$:
\begin{equation}
    F_{rep} = \lambda \cdot \left( \max\left(0, 1 - \frac{d}{d_{act}}\right) \right)^2 \cdot \frac{x_t - x_{obs}}{\|x_t - x_{obs}\|}
\end{equation}
This force is superposed onto the base velocity: $v_{final} = v_{base} + F_{rep}$.

\subsection{Evaluation and Compute Resources}

\subsubsection{Metrics Definition}
\begin{itemize}
    \item \textbf{Task Success:} A trajectory is considered successful if no collision and the final reward exceeds 85\% (Push-T) or if the object is successfully placed/lifted (Robomimic).
    \item \textbf{Inference Latency:} Measured as the average wall-clock time per control step. Note that for chunk-based baselines (DP, FP), the generation time is amortized over the execution chunk size ($H_{exec}=8$).
\end{itemize}

\subsubsection{Compute Hardware}
All experiments and latency measurements were conducted on a workstation with the following specifications:
\begin{itemize}
    \item \textbf{CPU:} AMD EPYC 7543 Processor.
    \item \textbf{GPU:} Single NVIDIA RTX A5000.
    \item \textbf{Framework:} PyTorch with CUDA acceleration.
\end{itemize}
\subsection{Complete Trade-off Plot}

\begin{figure}[t]
    \centering
    \includegraphics[width=0.45\textwidth]{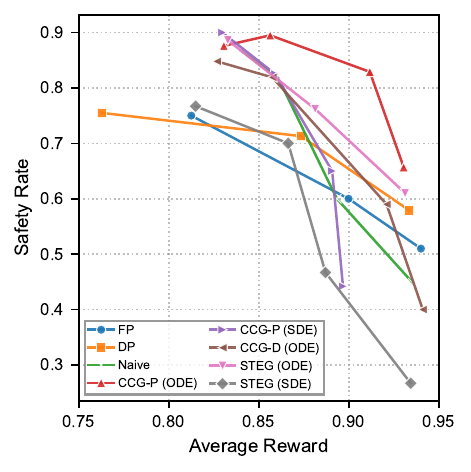}
    \caption{Comprehensive Safety-Reward trade-off (Average Reward vs. Safety Rate) across various baselines. We primarily report ODE variants in the main text, as the stochastic noise introduced by SDE inference was found to degrade performance.}
    \label{fig:tradeoffcomplete}
\end{figure}

Figure~\ref{fig:tradeoffcomplete} presents the complete Pareto frontier of safety versus task performance. We observe several additional key trends:

\paragraph{Impact of Stochasticity (SDE vs. ODE)}
A notable finding is the performance gap between ODE-based and SDE-based  inference. While SDEs theoretically offer better exploration, we find that in the context of high-frequency streaming control ($20$ Hz), the injected noise ($\gamma(t)\mathbf{z}$) disrupts the temporal coherence of the trajectory. The "jitter" introduced by the SDE solver makes fine manipulation tasks (e.g., precise block alignment in Push-T) difficult to stabilize, leading to lower Average Rewards. Consequently, the noise-free ODE formulation serves as a stronger backbone for reactive guidance. We therefore report the ODE-based inference results in the main text.

\paragraph{Baseline Comparison}

The energy-guided Diffusion Policy (DP) exhibits a sharp drop in reward as the guidance scale increases, without yielding significant gains in safety. We attribute this to the fundamental limitations of policies trained on proficient human datasets, which inherently lack the kinematic diversity required for flexible avoidance. While our method is subject to similar distributional constraints, the issue is significantly mitigated; the continuous nature of SSIP allows us to increase the guidance scale to effectively force a lateral detour, whereas similar scaling in DP destabilizes the generation process.

\end{document}